\documentclass[]{bytedance_seed}

\usepackage[toc,page,header]{appendix}
\usepackage{minitoc}
\usepackage{amsthm}
\usepackage{amssymb}
\usepackage{array}
\usepackage{longtable}
\usepackage{algorithm}
\usepackage{algorithmic}
\usepackage{url}

\titleformat*{\paragraph}{\sffamily\bfseries}

\usepackage{amsmath,amsfonts,bm}

\def\figref#1{figure~\ref{#1}}

\def\secref#1{section~\ref{#1}}

\def\eqref#1{equation~\ref{#1}}

\def\1{\bm{1}}

\def\vzero{{\bm{0}}}

\def\vb{{\bm{b}}}
\def\vc{{\bm{c}}}
\def\vd{{\bm{d}}}
\def\ve{{\bm{e}}}

\def\vg{{\bm{g}}}
\def\vh{{\bm{h}}}

\def\vm{{\bm{m}}}

\def\vp{{\bm{p}}}
\def\vq{{\bm{q}}}

\def\vs{{\bm{s}}}

\def\vu{{\bm{u}}}
\def\vv{{\bm{v}}}

\def\vx{{\bm{x}}}
\def\vy{{\bm{y}}}

\def\mA{{\bm{A}}}
\def\mB{{\bm{B}}}

\def\mI{{\bm{I}}}

\def\mL{{\bm{L}}}

\def\mV{{\bm{V}}}

\DeclareMathAlphabet{\mathsfit}{\encodingdefault}{\sfdefault}{m}{sl}
\SetMathAlphabet{\mathsfit}{bold}{\encodingdefault}{\sfdefault}{bx}{n}

\def\gJ{{\mathcal{J}}}

\def\gL{{\mathcal{L}}}

\def\gN{{\mathcal{N}}}

\def\gP{{\mathcal{P}}}

\def\gR{{\mathcal{R}}}

\newcommand{\E}{\mathbb{E}}

\newcommand{\R}{\mathbb{R}}

\newcommand{\KL}{D_{\mathrm{KL}}}

\definecolor{TRDBlue}{RGB}{0,92,165}
\definecolor{TRDOrange}{RGB}{190,80,0}
\definecolor{TRDGreen}{RGB}{0,120,90}
\definecolor{TRDPurple}{RGB}{125,70,145}

\newtheorem{proposition}{Proposition}
\newtheorem{theorem}{Theorem}
\newtheorem{corollary}{Corollary}
\newtheorem{assumption}{Assumption}
\theoremstyle{remark}
\newtheorem{remark}{Remark}

\newcommand{\sg}{\operatorname{sg}}
\newcommand{\veps}{{\bm{\epsilon}}}
\newcommand{\vxi}{{\bm{\xi}}}
\newcommand{\vdelta}{{\bm{\delta}}}

\newcommand{\vX}{{\bm{X}}}
\newcommand{\op}{\mathrm{op}}
\newcommand{\fitwidth}[1]{\resizebox{\ifdim\width>\linewidth\linewidth\else\width\fi}{!}{#1}}
\DeclareMathOperator{\Law}{Law}
\DeclareMathOperator{\rank}{rank}
\DeclareMathOperator{\Lip}{Lip}

\title{Trajectory-Regularized Distribution Distillation for One-Step Generation}
\author[1,2,*]{An Zhao}
\author[2]{Yixiang Zhou}
\author[2,\dagger]{Zejian Li}
\author[1]{Yi Guo}
\author[2]{Pei Chen}
\author[2]{Lingyun Sun}

\affiliation[1]{ByteDance Seed}
\affiliation[2]{Zhejiang University}

\contribution[*]{Work done during an internship at ByteDance Seed}
\contribution[\dagger]{Corresponding author}

\date{\today}
\correspondence{Zejian Li at \email{zejianlee@zju.edu.cn}}
\checkdata[Code]{\url{https://github.com/happyw1nd/TRD}}

\abstract{
Distilling flow matching models for one-step generation generally follows two paradigms.
Trajectory alignment supervises the student with the teacher's probability flow ODE, but its performance degrades under very small sampling budgets.
Distribution matching directly optimizes the generated distribution, but estimating its reverse Kullback-Leibler gradient typically requires a separate network to track the evolving student score.
We propose trajectory-regularized distribution distillation (TRD), which combines distribution-level and trajectory-level supervision through a momentum-mixture policy predicted in a single student forward pass.
This policy serves simultaneously as a one-step generator, an estimator of the student's marginal velocity that removes the need for a second trainable network in distribution matching, and a closed-form time-dependent velocity field supervised along the teacher ODE trajectory.
Experiments on SANA-1.6B and Stable Diffusion 3.5 show that TRD outperforms representative consistency and distribution matching baselines in generation quality and training efficiency for both one-step and few-step generation, with a trade-off in sample diversity.
}

\begin{document}
\maketitle

\section{Introduction}

Diffusion and flow matching models support high-resolution image synthesis \citep{rombach2022high,peebles2023scalable,esser2024scaling,xie2025sana}. Deterministic samplers for score-based diffusion and flow matching models numerically integrate a generative ODE, often requiring tens of network evaluations \citep{song2021score,karras2022elucidating,lipman2023flow}. Distillation seeks to reduce this cost to one or two evaluations \citep{salimans2022progressive,song2023consistency,yin2024onestep}; we organize prior approaches around two primary supervision objects.

Trajectory-oriented methods reduce long generative paths through recursive teacher-step compression \citep{salimans2022progressive}, consistency along solution curves \citep{song2023consistency,luo2023latent,kim2024consistency,lu2025simplifying}, path straightening \citep{liu2023flow,liu2024instaflow}, or interval-level policies \citep{chen2026piflow,yang2026arcflow}. Related direct-training objectives learn long transitions through improved consistency training, shortcuts or average velocities without requiring a fixed teacher \citep{song2024improved,frans2025one,geng2025mean}. For linear-shortcut distillation, \citet{yang2026arcflow} attribute quality loss at two to four evaluations to approximating a teacher path whose tangent direction changes over time.

\begin{figure}[t]
\centering
\includegraphics[width=\linewidth]{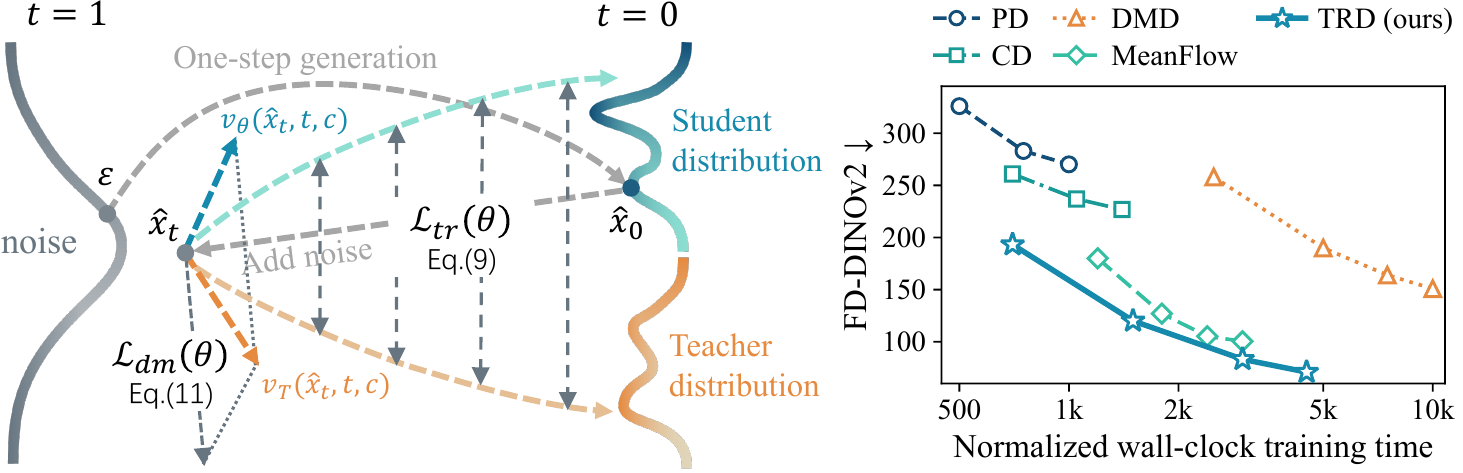}
\caption{Overview of TRD. Left: one student forward pass at a noised student state returns a momentum-mixture policy that supplies both the distribution matching direction of \eqref{eq:ldm} and the trajectory-grid velocities of \eqref{eq:ltr}. Right: one-step FD-DINOv2 against normalized wall-clock training time when distilling SANA-1.6B. Appendix~\ref{app:conventions} describes both panels.}
\label{fig:overview}
\end{figure}

Distribution matching instead aligns the student and teacher output laws through score differences \citep{luo2023diffinstruct,yin2024onestep,yin2024improved} or a Fisher-divergence objective \citep{zhou2024score}. It avoids prescribing a one-to-one teacher sampling path. The cited score-difference and Fisher methods typically train a second network to track the evolving student score; inaccurate tracking can destabilize training \citep{yin2024improved}, while the tracker adds memory and update overhead that becomes substantial at larger scales \citep{cheng2025twinflow,kim2026fsfdmd}.

TRD uses one policy-valued student output as a shared representation for generation, distribution matching, and trajectory alignment. It reuses the momentum-mixture parameterization of ArcFlow for closed-form curved generation \citep{yang2026arcflow}, and additionally uses the same output for student marginal velocity estimation and trajectory-grid prediction. Self flow matching, distribution matching, trajectory alignment, and distribution preservation jointly train this field. The controlled estimator ablation in Table~\ref{tab:fake} shows that sharing improves overall generation quality and teacher-distribution agreement relative to a DMD-style separate fake network, while avoiding its auxiliary optimizer state and gradient memory.

Our contributions are summarized as follows:
\begin{itemize}
    \item We use the momentum-mixture policy as a shared student representation for closed-form one-step generation, student marginal velocity estimation, and trajectory-grid velocity prediction.
    \item We formulate TRD to combine distribution-level and trajectory-level supervision through a shared self-estimator, removing the need for a separately trained fake model and its auxiliary optimizer state and gradient memory. Distribution preservation further stabilizes updates to the evolving student distribution.
    \item We establish ideal-state correctness for the combined objective, show that trajectory supervision of the shared start-time output rescales the reverse Kullback-Leibler update while preserving its direction, and provide finite-error bounds with complete proofs in \secref{sec:analysis} and Appendix~\ref{app:proofs}.
    \item We evaluate TRD on SANA-1.6B \citep{xie2025sana} and Stable Diffusion 3.5 Medium, whose architecture follows the multimodal rectified-flow transformer family of \citet{esser2024scaling}, through matched-NFE comparisons and controlled ablations that isolate the shared estimator, the objective terms and the policy capacity.
\end{itemize}

\section{Related work}
\label{sec:related}

\paragraph{Trajectory-oriented methods.}
Trajectory-oriented methods learn long transitions tied to sampling paths. Endpoint, progressive and bootstrapped distillation enlarge deterministic teacher transitions \citep{luhman2021knowledge,salimans2022progressive,gu2024boot}, while rectified flow and reflow straighten their couplings \citep{liu2023flow,liu2024instaflow}. Consistency and flow-map objectives enforce agreement across ODE points or intervals \citep{song2023consistency,song2024improved,luo2023latent,lu2025simplifying,kim2024consistency,boffi2025flowmap,sabour2025align}. Moment matching distils conditional moments \citep{salimans2024multistep}; MeanFlow and shortcut models directly learn average velocities or long transitions \citep{geng2025mean,frans2025one}; and policy-based methods distil interval dynamics \citep{chen2026piflow,yang2026arcflow}. TRD adds distribution matching to pointwise teacher-trajectory supervision through one policy.

\paragraph{Distribution distillation with a learned student-score model.}
Distribution methods compare noised generated and target laws. VSD uses target and learned variational scores \citep{wang2023prolificdreamer,nguyen2024swiftbrush}, and Diff-Instruct integrates divergences over time \citep{luo2023diffinstruct}. DMD combines a score-difference update with teacher-output regression \citep{yin2024onestep}; DMD2 removes the regression loss, stabilizes fake-score training with a two-time-scale update rule and adds adversarial training \citep{yin2024improved}. SiD and Guided SiD use Fisher-type objectives \citep{zhou2024score,zhou2025guided}, whereas SIM generalizes to a broader family of score-based divergences \citep{luo2024sim}. Each of these methods trains an auxiliary model to follow the evolving student score. TRD estimates the corresponding marginal velocity with its generation policy.

\paragraph{Hybrid and fake-score-free alternatives.}
Hybrid methods combine teacher-following with distributional or adversarial alignment. ADD, LADD, Diffusion2GAN, Flash Diffusion and SANA-Sprint mix score, paired-output or consistency supervision with discriminators or distribution losses \citep{sauer2024adversarial,sauer2024latent,kang2024distilling,chadebec2025flash,chen2025sanasprint}. SiDA adds adversarial supervision to score identity matching \citep{zhou2025adversarial}, ADM uses adversarial distribution matching \citep{lu2025adm}, and TDM matches trajectories as distributions \citep{luo2025tdm}. Other methods avoid a separately trained fake-score network through energy distance, contrastive energy estimation, endpoint pseudo-velocities, self-adversarial twin trajectories or network self-evaluation \citep{shao2025varflow,zhu2025simple,kim2026fsfdmd,cheng2025twinflow,yu2026selfevaluation}. TRD instead couples pointwise teacher velocities with reverse Kullback-Leibler matching through one policy, without a separate score model or discriminator.

\section{Preliminaries}
\label{sec:preliminaries}

\paragraph{Flow matching.}
We use the convention that noise sits at $t=1$ and data at $t=0$. For a data sample $\vx_0$ and $\veps\sim\gN(\vzero,\mI)$, the linear path is
\begin{equation}
\vx_t=(1-t)\vx_0+t\veps,
\qquad t\in[0,1].
\label{eq:path}
\end{equation}
Its conditional velocity is $\veps-\vx_0$. A teacher trained by conditional flow matching \citep{lipman2023flow,liu2023flow} predicts the marginal velocity $\vv_T(\vx,t,c)$ under a condition $c$, and sampling integrates $\dot\vx=\vv_T(\vx,t,c)$ from $t=1$ to $t=0$. Conditioning on $c$ is implicit in distribution notation. The core objective below uses the conditional teacher field directly; Appendix~\ref{app:qualifications} gives the classifier-free-guidance extension.

\paragraph{Distribution matching distillation.}
Let $G_\theta$ be a student generator with trainable parameters $\theta$. Given a condition $c$ and initial noise $\vx_1\sim\gN(\vzero,\mI)$, it produces the endpoint $\hat\vx_0=G_\theta(\vx_1,c)$. Draw $t\sim\rho$ on $[t_{\min},t_{\max}]\subset(0,1)$ and independent noise $\veps\sim\gN(\vzero,\mI)$ to form
\begin{equation}
\hat\vx_t=(1-t)\hat\vx_0+t\veps.
\label{eq:noised}
\end{equation}
We use $q_{\theta,t}$ and $p_{T,t}$ for the noised student and teacher laws, including their endpoint laws at $t=0$. Following integral divergence and distribution matching distillation updates \citep{luo2023diffinstruct,yin2024onestep}, we consider
\begin{equation}
\gJ_{\mathrm D}(\theta)
=\E_{c,t\sim\rho}\big[w_{\mathrm D}(t)\KL(q_{\theta,t}\Vert p_{T,t})\big],
\qquad w_{\mathrm D}(t)\ge0.
\label{eq:jd}
\end{equation}
Writing $\vs_{q_\theta,t}(\vx):=\nabla_\vx\log q_{\theta,t}(\vx)$ and $\vs_{p_T,t}(\vx):=\nabla_\vx\log p_{T,t}(\vx)$, its gradient under the regularity conditions in Appendix~\ref{app:proofs} is
\begin{equation}
\nabla_\theta\gJ_{\mathrm D}
=\E\big[w_{\mathrm D}(t)(J_\theta\hat\vx_t)^\top
\big(\vs_{q_\theta,t}(\hat\vx_t)-\vs_{p_T,t}(\hat\vx_t)\big)\big].
\label{eq:dm-score-gradient}
\end{equation}
The teacher supplies the target score, but the generator does not automatically supply the score of its own evolving output law. DMD therefore trains an auxiliary velocity model $\vv_\phi$ on generated samples with the flow-matching target $\veps-\hat\vx_0$, while the frozen teacher provides $\vv_T$. On the linear path~\eqref{eq:path}, DMD replaces the score difference in \eqref{eq:dm-score-gradient} by the velocity-form plug-in
\begin{equation}
\vs_{q_\theta,t}(\vx)-\vs_{p_T,t}(\vx)
\ \longrightarrow\
-\frac{1-t}{t}\big(\vv_\phi(\vx,t,c)-\vv_T(\vx,t,c)\big).
\label{eq:dmd-velocity-substitution}
\end{equation}
The generator is held fixed when $\vv_\phi$ is trained. We set $\lambda(t)=w_{\mathrm D}(t)(1-t)/t$ below; the additional regression and adversarial terms of DMD and DMD2 are discussed in \secref{sec:related}.

\paragraph{ArcFlow.}
ArcFlow replaces a constant-velocity shortcut with a momentum-mixture velocity parameterization that represents velocity evolution over an entire sampling interval \citep{yang2026arcflow}. Its backbone feeds three output projections that produce $K$ base velocities, softmax-normalized gate weights and positive momentum factors, which we call momentum rates below. One network evaluation at a fixed input $(\vx_t,t,c)$ returns the momentum-mixture parameters
\begin{equation}
\begin{aligned}
\gP_\theta(\vx_t,t,c)
&=\{\vv_k,\pi_k,\gamma_k\}_{k=1}^{K},
\qquad \pi_k>0,\quad \sum_k\pi_k=1,\quad \gamma_k>0,\\
\vv_{t,s}^{\theta}
&:=\tilde\vv_\theta\big(s;\gP_\theta(\vx_t,t,c)\big)
=\sum_{k=1}^{K}\pi_k\vv_k\gamma_k^{1-s},
\qquad 0\le s\le t .
\end{aligned}
\label{eq:policy}
\end{equation}
In $\vv_{t,s}^{\theta}$, $t$ denotes the time of the input state $\vx_t$, while $s\in[0,t]$ denotes the time at which the resulting velocity is evaluated. The starting state and condition are implicit in $\vv_{t,s}^{\theta}$. With the predicted coefficients fixed, $s\mapsto\vv_{t,s}^{\theta}$ defines the student's velocity parameterization over the interval $[0,t]$. Its exact integral defines the endpoint generator
\[
G_\theta(\vx_t,t,c)
=\vx_t-\int_0^t\vv_{t,s}^{\theta}\,\mathrm ds
=\vx_t-\sum_{k=1}^{K}\pi_k\vv_k\gamma_k^{1-t}
\int_0^t\gamma_k^u\,\mathrm du.
\]
We abbreviate $G_\theta(\vx_1,1,c)$ as $G_\theta(\vx_1,c)$, and \eqref{eq:rollout} in Appendix~\ref{app:policyfield} states the rollout to an arbitrary earlier time.

\section{Trajectory-regularized distribution distillation}
\label{sec:method}

Trajectory-regularized distribution distillation (TRD) combines distribution matching and teacher-trajectory alignment through a shared ArcFlow parameterization.
Distribution matching typically trains an auxiliary model to track the marginal velocity of the student's generated distribution; because an ArcFlow student already outputs a time-dependent velocity, TRD trains the student itself to provide this estimate.
We first combine the two signals while retaining the auxiliary model (\secref{sec:traj}), and then replace it with self flow matching (\secref{sec:dmd}).

\subsection{Distribution matching with trajectory regularization}
\label{sec:traj}

\paragraph{Distribution matching with an auxiliary estimator.}
We first apply the conventional DMD construction of \secref{sec:preliminaries} to the ArcFlow generator $G_\theta$, retaining $\vv_\phi$ as the auxiliary velocity model. Using the velocity-form substitution in \eqref{eq:dmd-velocity-substitution}, the gradient in \eqref{eq:dm-score-gradient} is implemented by
\begin{equation}
    \label{eq:first_ldm}
    \gL_{\mathrm{dm}}(\theta;\phi)
    =\E\Big[\Big\langle
        \sg\big[-\lambda(t)(\vv_\phi(\hat\vx_t,t,c)-\vv_T(\hat\vx_t,t,c))\big],
        \hat\vx_t\Big\rangle\Big].
\end{equation}

We write $\sg[\cdot]$ for stop-gradient. Because the bracketed velocity difference is detached, automatic differentiation acts only through the path $\hat\vx_t\rightarrow\hat\vx_0\rightarrow\theta$. The resulting gradient agrees with \eqref{eq:dm-score-gradient} when $\vv_\phi$ and $\vv_T$ recover their corresponding marginal velocities; the scalar value of the surrogate is not the KL divergence itself.

\paragraph{Trajectory alignment from the re-noised state.}
The state $\hat\vx_t$ is obtained by adding independent noise to the generated endpoint $\hat\vx_0$; it need not lie on the original generation trajectory. Given a decreasing timestep grid $\mathcal S_t=\{t,s_1,\ldots,s_M\}$ with $t>s_1>\cdots>s_M\ge0$, let $\hat\vx_{t,s}^{T}$ be the teacher state reached at time $s$, starting from $\hat\vx_t$ at $t$. For consecutive grid times $s>s'$, explicit Euler gives
\begin{equation}
\hat\vx_{t,t}^{T}=\hat\vx_t,\qquad \vv_{t,s}^{T}=\vv_T(\hat\vx_{t,s}^{T},s,c),\ s\in\mathcal S_t,\qquad \hat\vx_{t,s'}^{T}=\hat\vx_{t,s}^{T}-(s-s')\vv_{t,s}^{T}.
\label{eq:teacherroll}
\end{equation}
The entire teacher rollout is computed without gradients. The student is evaluated once at $(\sg[\hat\vx_t],t,c)$, giving the coefficients of \eqref{eq:policy}. Thus, evaluating $\vv_{t,s}^{\theta}$ at every grid time requires no additional student evaluations. The trajectory alignment loss is defined as
\begin{equation}
\gL_{\mathrm{tr}}(\theta)
=\E\bigg[\frac{1}{M+1}\sum_{s\in\mathcal S_t}
\big\|\vv_{t,s}^{\theta}-\sg[\vv_{t,s}^{T}]\big\|_2^2\bigg].
\label{eq:ltr}
\end{equation}
This loss teaches the student's velocity parameterization to represent the teacher's velocity field over the whole grid. It compares velocities, but not student endpoints or student states along the grid. Proposition~\ref{prop:grid} and Theorem~\ref{thm:policy} in Appendix~\ref{app:proofs} give the corresponding expressivity and finite-grid error results.

\subsection{Replacing the auxiliary model with self flow matching}
\label{sec:dmd}

We observe that the velocity parameterization already produces a start-time instantaneous velocity $\vv_{t,t}^{\theta}$, which can estimate the marginal velocity of the student's generated law.
We train this existing output to estimate the marginal velocity through a self flow matching loss, removing the separate auxiliary model:
\begin{equation}
\gL_{\mathrm{fm}}(\theta)
=\E\Big[\big\|\vv_{t,t}^{\theta}
-\big(\veps-\sg[\hat\vx_0]\big)\big\|_2^2\Big].
\label{eq:lfm}
\end{equation}

Here the student is evaluated at the detached state $\sg[\hat\vx_t]$, as in trajectory alignment; gradients update the parameters producing $\vv_{t,t}^{\theta}$, while the input $\hat\vx_t$ and endpoint target $\hat\vx_0$ remain detached. For a fixed student law, the minimizer of this regression is $\E[\veps-\hat\vx_0\mid\hat\vx_t,t,c]$, the student's marginal velocity at $\hat\vx_t$; Proposition~\ref{prop:selffm} gives the formal statement.

Replacing $\vv_\phi$ by this shared estimate gives the TRD distribution update:
\begin{equation}
\gL_{\mathrm{dm}}(\theta)
=\E\Big[\big\langle
\sg\big[-\lambda(t)(\vv_{t,t}^{\theta}-\vv_{t,t}^{T})\big],
\hat\vx_t\big\rangle\Big].
\label{eq:ldm}
\end{equation}

Here $\langle\cdot,\cdot\rangle$ sums over all non-batch axes. Compared with the surrogate in \eqref{eq:first_ldm}, \eqref{eq:ldm} replaces only the auxiliary estimate $\vv_\phi(\hat\vx_t,t,c)$ with the shared prediction $\vv_{t,t}^{\theta}$; the stop-gradient construction and endpoint gradient path remain unchanged. When $\vv_{t,t}^{\theta}$ and $\vv_{t,t}^{T}$ recover the corresponding marginal velocities, Theorem~\ref{thm:dmd} shows that its automatic-differentiation gradient equals $\nabla_\theta\gJ_{\mathrm D}$. Distribution matching and self flow matching share the start-time prediction $\vv_{t,t}^{\theta}$, while trajectory alignment evaluates the same velocity parameterization over $s\in\mathcal S_t$.

\paragraph{Interaction with trajectory alignment.}
Because $t\in\mathcal S_t$, \eqref{eq:ltr} pulls $\vv_{t,t}^{\theta}$ toward the teacher velocity while \eqref{eq:lfm} pulls it toward the student marginal velocity, and \eqref{eq:ldm} uses their difference. Proposition~\ref{prop:compromise} in Appendix~\ref{app:proofdist} shows that, when the start-time output is otherwise unconstrained, the joint minimizer is a convex combination of the two targets. The resulting update is the exact reverse-KL update scaled by a factor $\alpha\ge\lambda_{\mathrm{fm}}/(\lambda_{\mathrm{fm}}+\lambda_{\mathrm{tr}})$, which exceeds $0.9$ for the weights in Table~\ref{tab:hparams}; trajectory alignment therefore reduces the distribution step size and leaves its direction and zeros unchanged.

\subsection{Stabilization and overall objective}
\label{sec:objective}

\paragraph{Distribution preservation.}
We stabilize distribution matching with a moving-average copy. Let $\bar\theta$ be the exponential moving average of $\theta$ at rate $\mu\in[0,1)$. Distribution preservation matches paired endpoints generated from the same initial noise:
\begin{equation}
\gL_{\mathrm{dp}}(\theta)
=\E_{\vx_1,c}\Big[
\big\|G_\theta(\vx_1,c)-\sg[G_{\bar\theta}(\vx_1,c)]\big\|_2^2\Big].
\label{eq:ldp}
\end{equation}
This loss keeps the current generator map close to its moving-average reference and vanishes when $\theta=\bar\theta$. It constrains the original generated endpoint $\hat\vx_0$, not an endpoint of the rollout from $\hat\vx_t$.

\paragraph{Overall objective.}
Trajectory alignment follows the teacher path, while distribution alignment combines reverse Kullback-Leibler matching, self flow matching and distribution preservation.
\begin{equation}
\begin{aligned}
\gL_{\mathrm{TRD}}(\theta)
={}&
\underbrace{
\lambda_{\mathrm{tr}}\gL_{\mathrm{tr}}(\theta)
}_{\text{trajectory alignment}\;(\ref{eq:ltr})}
+
\underbrace{
\lambda_{\mathrm{dm}}\gL_{\mathrm{dm}}(\theta)
}_{\text{reverse-KL matching}\;(\ref{eq:ldm})}
&+
\underbrace{
\lambda_{\mathrm{fm}}\gL_{\mathrm{fm}}(\theta)
}_{\text{self flow matching}\;(\ref{eq:lfm})}
+
\underbrace{
\lambda_{\mathrm{dp}}\gL_{\mathrm{dp}}(\theta)
}_{\text{distribution preservation}\;(\ref{eq:ldp})}.
\end{aligned}
\label{eq:total}
\end{equation}
One ArcFlow parameterization serves three roles: its analytic integral generates endpoints, its start-time velocity estimates the student's marginal velocity, and its velocities at earlier times provide predictions for trajectory alignment.
The student is initialized from the teacher. Its final $L$ transformer blocks are copied into $K$ momentum branches, where $L$ denotes the duplicated depth and each branch predicts $(\vv_k,\pi_k,\gamma_k)$ in \eqref{eq:policy}. Only these branches and the low-rank backbone adapters are trained \citep{hu2022lora}.

\subsection{Training procedure}
\label{sec:algo}

Algorithm~\ref{alg:trd} summarizes one TRD training iteration. We first generate and re-noise a student endpoint, evaluate the student at the detached noised state to obtain its velocity parameterization, and roll out the frozen teacher from the same state. This requires two student forward passes: one for generation and one for the shared velocity predictions. The teacher trajectory and moving-average endpoint define the remaining targets. All four losses are formed before one joint student update, after which the moving average is updated.

\begin{algorithm}[H]
\caption{One TRD training iteration for one-step distillation}
\label{alg:trd}
\begin{algorithmic}[1]
\REQUIRE student $\theta$, moving average $\bar\theta$ with rate $\mu$, frozen teacher $\vv_T$, time distribution $\rho$ on $[t_{\min},t_{\max}]$, grid $\mathcal S_t=\{t,s_1,\ldots,s_M\}$ with $t>s_1>\cdots>s_M\ge0$, loss weights $\lambda_{\mathrm{tr}},\lambda_{\mathrm{dm}},\lambda_{\mathrm{fm}},\lambda_{\mathrm{dp}}$ of \eqref{eq:total} and time weight $\lambda(t)$
\STATE sample $c$, initial noise $\vx_1\sim\gN(\vzero,\mI)$, $t\sim\rho$ and independent $\veps\sim\gN(\vzero,\mI)$
\STATE {\bfseries generate:} evaluate the student at $(\vx_1,1,c)$ and integrate the output velocity to $\hat\vx_0=G_\theta(\vx_1,c)$
\STATE {\bfseries re-noise:} form $\hat\vx_t=(1-t)\hat\vx_0+t\veps$, retaining its gradient path to $\theta$
\STATE {\bfseries predict velocities:} evaluate the student at $(\sg[\hat\vx_t],t,c)$; obtain $\{\vv_{t,s}^{\theta}:s\in\mathcal S_t\}$ from \eqref{eq:policy}, without further student evaluations
\STATE {\bfseries teacher rollout, without gradients:} initialize $\hat\vx_{t,t}^{T}=\hat\vx_t$; for decreasing $s\in\mathcal S_t$, evaluate $\vv_{t,s}^{T}$ and, if a next grid time $s'$ exists, advance to $\hat\vx_{t,s'}^{T}$ by \eqref{eq:teacherroll}
\STATE {\bfseries reference endpoint, without gradients:} evaluate $G_{\bar\theta}(\vx_1,c)$ on the same initial noise
\STATE {\bfseries form the losses:} use the grid velocities for \eqref{eq:ltr}, $\vv_{t,t}^{\theta}$ and $\veps-\sg[\hat\vx_0]$ for \eqref{eq:lfm}, the detached start-time velocity difference and $\hat\vx_t$ for \eqref{eq:ldm}, and the paired endpoints for \eqref{eq:ldp}
\STATE {\bfseries update:} backpropagate \eqref{eq:total}, take one optimizer step on $\theta$, then set $\bar\theta\leftarrow\mu\bar\theta+(1-\mu)\theta$
\end{algorithmic}
\end{algorithm}

\paragraph{Few-step extension.}
Inspired by DMD2's backward simulation \citep{yin2024improved}, few-step TRD uses random-length consistency sampling. Fix $1=a_0>\cdots>a_R=0$ and sample $r\sim\operatorname{Unif}\{1,\ldots,R\}$. Starting from $\vx_{a_0}\sim\gN(\vzero,\mI)$, run $r-1$ detached stages that predict an endpoint and independently re-noise it to the next preset time; only $\hat\vx_0=G_\theta(\vx_{a_{r-1}},a_{r-1},c)$ retains gradients. Then sample $t\sim\rho_r$ with $\operatorname{supp}(\rho_r)\subseteq[a_r,a_{r-1}]$, form \eqref{eq:noised} and reuse the four losses. Distribution preservation compares student and moving-average endpoints from the same $\vx_{a_{r-1}}$. The stage times $a_i$ differ from the trajectory-supervision times $s\in\mathcal S_t$; Algorithm~\ref{alg:fewstep} in Appendix~\ref{app:impl} gives full details.

\subsection{Training target analysis}
\label{sec:analysis}

The four terms of \eqref{eq:total} supervise different objects, so we check that they admit a common ideal target. Assume exact expectations, an exact unguided teacher, $\lambda(t)>0$ on a set of positive $\rho$-measure, and a student expressive enough to represent the teacher endpoint law, its own marginal velocity and the teacher trajectory.

Let $\vv_{q_\theta,t}$ and $\vv_{p_T,t}$ denote the true marginal velocity fields of $q_{\theta,t}$ and $p_{T,t}$, respectively.

\begin{proposition}[Shared ideal target]
\label{prop:ideal}
Under the assumptions above, the following statements hold:
(i) the population minimizer of \eqref{eq:lfm} is the student
marginal velocity $\vv_{q_\theta,t}$;
(ii) with that velocity and the exact teacher velocity in place,
the automatic-differentiation gradient of \eqref{eq:ldm} equals
the gradient of
\[
\gJ_{\mathrm D}(\theta)
=\E_{c,t\sim\rho}\big[
w_{\mathrm D}(t)\KL(q_{\theta,t}\Vert p_{T,t})\big],
\qquad
w_{\mathrm D}(t)=\frac{t\lambda(t)}{1-t}\ge0,
\]
and every global minimizer of $\gJ_{\mathrm D}$ satisfies
$q_{\theta,0}=p_{T,0}$;
(iii) if the student attains $q_{\theta,0}=p_{T,0}$, predicts its
own marginal velocity, and has zero continuous-time trajectory
residual, and if $\bar\theta=\theta$, then the expected update
induced by each of the four terms in \eqref{eq:total} is zero.
\end{proposition}

Proposition~\ref{prop:ideal} shows that replacing the auxiliary model with self flow matching preserves the ideal distribution target and that all four expected updates share a common zero. This common zero does not make every stationary point of the shared-parameter update correct, and it does not imply global convergence. Appendix~\ref{app:proofs} gives the regularity conditions and finite-error bounds, and Appendix~\ref{app:qualifications} treats guidance, gradient bias and local stability.

\section{Experiments}
\label{sec:experiments}

\subsection{Setup}
\label{sec:setup}

\paragraph{Teachers, prompts and metrics.}
We distil SANA-1.6B \citep{xie2025sana} and Stable Diffusion 3.5 Medium, from the multimodal rectified-flow transformer family \citep{esser2024scaling}, at $1024\times1024$. Students initialize from their teachers and train only the policy heads and backbone adapters; TRD uses $K=4$ and $L=1$ unless stated otherwise. All runs use disjoint splits of a publicly available dataset \citep{schuhmann2022laion}, including 2{,}500 test prompts. We evaluate teacher-relative fidelity and coverage, diversity, preference, visual appeal and text alignment. Appendix~\ref{app:impl} gives the full configurations and protocols.

\paragraph{Baselines.} Table~\ref{tab:sana} groups baselines by trajectory-oriented \citep{salimans2022progressive,song2023consistency,song2024improved,lu2025simplifying,yang2026arcflow,geng2025mean} or distribution-oriented supervision \citep{yin2024onestep,yin2024improved,zhou2024score,kim2026fsfdmd,cheng2025twinflow}; TRD combines both. For comparability, we adapt the teacher-free MeanFlow and TwinFlow regimes to teacher initialization and the shared data, resolution and target-NFE protocol.

\subsection{One-step and few-step distillation}
\label{sec:main}

Table~\ref{tab:sana} reports one-step and few-step distillation results for SANA-1.6B, with the multi-step teacher as an upper reference; the matched comparison on Stable Diffusion 3.5 Medium is deferred to Table~\ref{tab:sd35} in Appendix~\ref{app:results}. Figures~\ref{fig:qualitative-sana} and~\ref{fig:qualitative-sd35} in Appendix~\ref{app:results} compare one-step and few-step samples from PD, CD, MeanFlow, DMD and TRD for the two teachers. Figures~\ref{fig:additional-sana} and~\ref{fig:additional-sd35} provide larger sets of one-step TRD samples.
At both NFEs, TRD leads the preference, aesthetic and text-alignment metrics and attains the lowest FD-DINOv2, and at two steps several of its preference and aesthetic scores exceed the multi-step teacher. It favors quality over diversity more strongly than DMD and related distribution-matching methods: its DINO Diversity falls behind the most diverse baselines at both NFEs, and at two steps its Recall-DINOv2 also falls behind DMD and DMD2, which cover more of the teacher manifold. Trajectory-oriented methods such as MeanFlow retain higher diversity, but lag on preference scores and Fr\'echet distance. Stable Diffusion 3.5 Medium shows the same pattern, except that DMD attains a slightly lower FD-DINOv2 at two steps.

\begin{table}[tp]
\caption{One-step and few-step results at $1024\times1024$ for students distilled from SANA-1.6B, with the multi-step teacher as a reference. Appendix~\ref{app:impl} defines the metrics and Appendix~\ref{app:conventions} gives the table conventions.}
\label{tab:sana}
\begin{center}
\footnotesize
\setlength{\tabcolsep}{4pt}
\fitwidth{%
\begin{tabular}{lcccccccccc}
\toprule
Method & NFE & Extra models & FD-DINOv2 $\downarrow$ & Pick $\uparrow$ & HPS $\uparrow$ & ImgRwd $\uparrow$ & Aesth $\uparrow$ & CLIP $\uparrow$ & Recall $\uparrow$ & Diversity $\uparrow$ \\
\midrule
\textcolor{gray}{SANA-1.6B teacher} & \textcolor{gray}{$20\times2$} & \textcolor{gray}{--} & \textcolor{gray}{--} & \textcolor{gray}{0.1971} & \textcolor{gray}{0.2360} & \textcolor{gray}{0.5144} & \textcolor{gray}{6.1879} & \textcolor{gray}{28.7036} & \textcolor{gray}{--} & \textcolor{gray}{0.4100} \\
\midrule
\multicolumn{11}{l}{\textit{Trajectory-oriented objectives}} \\
PD & 1 & 0 & 260.81 & 0.1764 & 0.1952 & -1.3441 & 5.6703 & 24.9226 & 0.0338 & 0.3362 \\
CD & 1 & 0 & 226.80 & 0.1814 & 0.2058 & -0.8208 & 5.7909 & 25.9214 & 0.0501 & 0.3181 \\
sCD & 1 & 0 & 202.58 & 0.1829 & 0.2103 & -0.6012 & 5.7925 & 27.0117 & 0.0721 & 0.3290 \\
ArcFlow & 1 & 0 & 193.68 & 0.1816 & 0.2078 & -0.4764 & 5.8057 & 26.8843 & 0.0755 & 0.3411 \\
MeanFlow & 1 & 0 & \underline{97.21} & 0.1829 & 0.2109 & -0.1739 & 5.9320 & 27.1596 & \underline{0.3031} & \textbf{0.4366} \\
\multicolumn{11}{l}{\textit{Distribution-oriented objectives}} \\
DMD & 1 & 1 & 150.11 & 0.1840 & 0.2129 & -0.3398 & 5.8483 & 28.1567 & 0.0854 & 0.3250 \\
DMD2 & 1 & 2 & 112.47 & \underline{0.1894} & 0.2192 & 0.1247 & 6.0468 & \underline{28.1960} & 0.1740 & \underline{0.3499} \\
SiD & 1 & 1 & 134.82 & 0.1877 & 0.2168 & -0.0185 & 5.9812 & 27.4055 & 0.1478 & 0.3321 \\
FSF-DMD & 1 & 0 & 128.53 & 0.1873 & \underline{0.2197} & \underline{0.2259} & \underline{6.0810} & 27.9147 & 0.0953 & 0.3142 \\
TwinFlow & 1 & 0 & 130.29 & 0.1861 & 0.2139 & 0.0526 & 5.9782 & 27.5482 & 0.0600 & 0.3233 \\
\multicolumn{11}{l}{\textit{Combined trajectory and distribution objective}} \\
TRD (ours) & 1 & 0 & \textbf{70.34} & \textbf{0.1915} & \textbf{0.2223} & \textbf{0.3888} & \textbf{6.1169} & \textbf{28.3647} & \textbf{0.3162} & 0.3204 \\
\midrule
\multicolumn{11}{l}{\textit{Trajectory-oriented objectives}} \\
PD & 2 & 0 & 91.50 & 0.1833 & 0.2128 & -0.3924 & 5.8369 & 26.7599 & 0.4250 & \underline{0.4686} \\
CD & 2 & 0 & 177.65 & 0.1779 & 0.2133 & -0.6014 & 5.6130 & 26.0449 & 0.1800 & 0.4027 \\
sCD & 2 & 0 & 137.31 & 0.1891 & 0.2124 & 0.1128 & 5.7723 & 26.6751 & 0.3691 & 0.4254 \\
ArcFlow & 2 & 0 & 138.74 & 0.1862 & 0.2091 & 0.0919 & 6.1177 & 26.6285 & 0.2825 & \textbf{0.4715} \\
MeanFlow & 2 & 0 & 83.89 & 0.1917 & 0.2151 & 0.1388 & 6.2768 & 27.0492 & 0.4934 & 0.4287 \\
\multicolumn{11}{l}{\textit{Distribution-oriented objectives}} \\
DMD & 2 & 1 & \underline{29.05} & \underline{0.1981} & 0.2339 & 0.4927 & \underline{6.3019} & \underline{28.4341} & \textbf{0.8263} & 0.3978 \\
DMD2 & 2 & 2 & 35.92 & 0.1975 & \underline{0.2346} & 0.4861 & 6.2987 & 28.0849 & \underline{0.7251} & 0.3732 \\
SiD & 2 & 1 & 47.58 & 0.1974 & 0.2325 & \underline{0.5037} & 6.2126 & 27.8064 & 0.4527 & 0.3929 \\
FSF-DMD & 2 & 0 & 39.41 & 0.1927 & 0.2308 & 0.4584 & 6.1869 & 28.0417 & 0.3308 & 0.3877 \\
TwinFlow & 2 & 0 & 36.57 & 0.1941 & 0.2285 & 0.4273 & 6.0078 & 28.0961 & 0.4068 & 0.3640 \\
\multicolumn{11}{l}{\textit{Combined trajectory and distribution objective}} \\
TRD (ours) & 2 & 0 & \textbf{28.49} & \textbf{0.1993} & \textbf{0.2353} & \textbf{0.6746} & \textbf{6.3105} & \textbf{28.4355} & 0.7151 & 0.3380 \\
\bottomrule
\end{tabular}
}
\end{center}
\end{table}

\subsection{Training cost and time-to-quality trade-offs}
\label{sec:cost}

A practical motivation for shared self-estimation is to remove the persistent parameters, gradients and optimizer state of the auxiliary fake network used by DMD-style methods. We therefore compare the number of auxiliary networks, trainable parameters, measured peak memory, normalized per-iteration time and normalized time to a common FD-DINOv2 target. Table~\ref{tab:cost} reports these measurements for full-scale one-step training. Times are normalized to one DMD training step; Appendix~\ref{app:cost} accounts for the underlying operations and persistent state. TRD uses $K=4$ momentum modes and $L=1$ duplicated final block. The right panel of \figref{fig:overview} plots FD-DINOv2 against normalized training time.

\begin{table}[ht]
\caption{Measured memory and time cost of one-step distillation from SANA-1.6B under full-scale training. Appendix~\ref{app:conventions} defines the reported columns.}
\label{tab:cost}
\begin{center}
\footnotesize
\setlength{\tabcolsep}{4pt}
\fitwidth{%
\begin{tabular}{lccccc}
\toprule
Method & Aux.\ nets & Trainable params & Peak mem.\ (GB) $\downarrow$ & Norm.\ time / iter $\downarrow$ & Norm.\ time to target $\downarrow$ \\
\midrule
MeanFlow & 0 & 1.6B & 40 & 1.2 & 2400 \\
CD & 0 & 1.6B & 40 & 0.7 & 1400 \\
DMD & 1 & 3.2B & 75 & 1.0 & 10000 \\
DMD2 & 2 & 3.3B & 80 & 1.2 & 12000 \\
TRD (ours) & 0 & 1.9B & 45 & 1.5 & 4500 \\
\bottomrule
\end{tabular}
}
\end{center}
\end{table}

Relative to DMD and DMD2, TRD removes one or two auxiliary networks, reducing trainable parameters and measured peak memory. Teacher-trajectory supervision makes each iteration more expensive, but TRD reaches the common target in substantially less normalized time. Within DMD-style distillation, TRD therefore trades per-iteration computation for less persistent training state and faster measured time to target. MeanFlow and CD remain cheaper in these measurements, so TRD is not the lowest-cost method across all distillation families.

\subsection{Dissecting the TRD design}
\label{sec:ablation}

We examine three questions about the TRD design: whether shared self-estimation improves on a separate student-velocity estimator, how the objective terms complement one another and how the capacity of the velocity parameterization affects performance. Unless stated otherwise, the comparisons use one-step SANA-1.6B distillation; Appendices~\ref{app:impl} and~\ref{app:results} give the full settings and results.

\paragraph{Shared generation and velocity estimation.}
Table~\ref{tab:fake} replaces the shared velocity estimate with a separately trained DMD-style fake network under otherwise matched settings. The shared parameterization improves FD-DINOv2, preference metrics and teacher coverage, whereas the separate estimator retains slightly greater diversity. Shared self-estimation therefore provides a measured quality benefit in addition to removing the auxiliary trainable model and its associated state.

\paragraph{Complementary objective roles.}
Under matched settings, Table~\ref{tab:loss} shows that removing trajectory alignment degrades fidelity, preference and coverage while preserving similar diversity, demonstrating benefits beyond the distribution objectives. Jointly removing self flow matching and distribution matching broadly degrades performance, although this comparison does not isolate their individual effects. Removing distribution preservation collapses training, indicating that it is critical for stability in this setting.

\paragraph{Velocity-parameterization capacity.}
Table~\ref{tab:policy} varies the number of momentum modes $K$ and the duplicated depth $L$. At $L=1$, increasing $K$ from one to two improves every reported metric, but four or eight modes reverse this trend despite the larger representational family of Proposition~\ref{prop:grid}; \figref{fig:momentum} rules out simple collapse of the gates and rates at $t=1$ but does not explain this reversal. At fixed $K$, increasing $L$ consistently improves quality and coverage with nearly unchanged diversity. Duplicated depth therefore gives a predictable trade-off between quality and cost, whereas $K$ should be tuned.

\subsection{Understanding TRD's performance gains}
\label{sec:mechanism}

\begin{figure}[t]
\centering
\includegraphics[width=\linewidth]{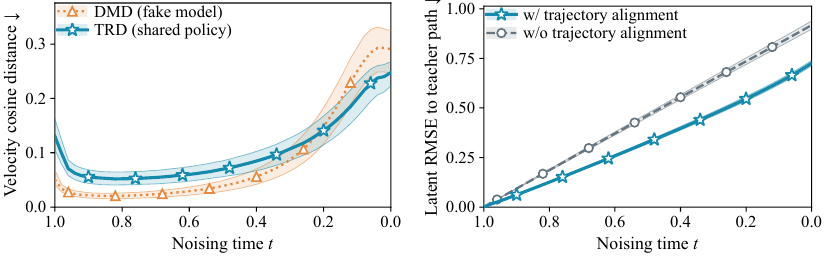}
\caption{Mechanism analysis at the final checkpoint. Left: velocity cosine distance against noising time for DMD and TRD. Right: discrepancy from the teacher denoising path for TRD with and without trajectory alignment. Appendix~\ref{app:conventions} describes both panels.}
\label{fig:mechanism}
\end{figure}

\figref{fig:mechanism} examines two possible sources of TRD's gains on final Stable Diffusion 3.5 Medium checkpoints; Appendix~\ref{app:conventions} gives the protocol.

\paragraph{Student velocity estimation.}
DMD uses a separately trained fake velocity estimator, while TRD reads the estimate from its generation policy. The left panel shows that TRD has weaker directional agreement at high noise but becomes more accurate near the data endpoint. Because this estimate supplies the distribution-matching direction in \eqref{eq:ldm}, the low-noise advantage is consistent with TRD's stronger one-step quality in Table~\ref{tab:sd35}. This diagnostic is correlational and does not establish causality.

\paragraph{Trajectory alignment.}
The right panel compares matched TRD variants with and without $\gL_{\mathrm{tr}}$. Trajectory alignment lowers the discrepancy from the teacher path at every noninitial grid point, although each training grid covers only a short segment below its resampled start time (Table~\ref{tab:hparams}).

\section{Conclusion}
\label{sec:conclusion}

We introduced trajectory-regularized distribution distillation, whose single momentum-mixture policy supports closed-form generation, student marginal velocity estimation and teacher-trajectory prediction without a separately trained fake model. Our analysis identifies a common ideal target for the supervision signals, shows that trajectory supervision of the shared estimator preserves the direction of the distribution update, and bounds their finite-error complementarity. Experiments on two text-to-image models support the quality benefits of this design and its efficiency relative to fake-model distribution matching, while revealing a trade-off in teacher-relative coverage and diversity. The evaluation covers two public image models, measures fidelity against teacher samples, reports no run-to-run variability and uses feature-space diversity metrics that miss some forms of conditional diversity; why training collapses without distribution preservation also remains open. Future work should test broader model families, real-data benchmarks and guidance settings.

\subsection*{Reproducibility statement}

The method is specified by \eqref{eq:total}, \eqref{eq:policy} through \eqref{eq:ldp}, and Algorithms~\ref{alg:trd} and~\ref{alg:fewstep}. Appendix~\ref{app:impl} records the teacher checkpoint identifiers, policy configuration, objective weights and schedules, trajectory grid and solver, optimizer settings, prompt split size and seed, shared image-seed schedule, metric definitions and model families. Appendix~\ref{app:proofs} formalizes the ideal argument of \secref{sec:analysis} under Assumption~\ref{ass:reg} and gives complete proofs of Propositions~\ref{prop:grid}, \ref{prop:selffm}, \ref{prop:compromise} and \ref{prop:dp}, Theorems~\ref{thm:dmd}, \ref{thm:endpoint} and \ref{thm:policy}, and Corollary~\ref{cor:joint}. Appendix~\ref{app:qualifications} gives the guidance extension, estimation-error analysis and convergence qualifications. 

\subsection*{Ethics statement}

TRD reduces the inference cost of pretrained text-to-image generators and requires no additional target images; the image experiments use prompts from a publicly available dataset during distillation. It therefore inherits the properties of its teachers, including any demographic or cultural bias in their training corpora and their capacity to produce misleading or harmful image content. Lowering inference cost can increase the rate at which such content is produced. The distillation objective in \eqref{eq:ldm} is mode-seeking, so a distilled student can narrow the output distribution relative to its teacher in ways that aggregate quality metrics do not reveal. We recommend that any deployment evaluate diversity and safety on its own input distribution. The text-to-image teachers, evaluation datasets, benchmarks and metric models are publicly released and are used under their published licenses.

\subsection*{AI use statement}

We used a large language model as a writing assistant for grammatical correction, structural revision and improving clarity in the introduction, method, experimental description, ideal-state correctness argument and appendix, and as a coding assistant for routine implementation of evaluation scripts and plotting utilities. Its mathematical role was limited to reorganizing and rephrasing existing author-provided definitions, claims and derivations. We did not use generative AI to originate research ideas, derive the method or theoretical claims, design experiments, or generate, alter or select any experimental result. All text produced with AI assistance was read and edited by at least two authors, every mathematical edit was checked against the original definitions and derivations, all code produced with AI assistance was reviewed line by line and validated against reference implementations of the corresponding metrics, and every reference was checked against the authoritative paper or technical report available for the cited work. The authors take full responsibility for the final content of this work, including its text, claims, figures, tables and released artifacts.

\clearpage
\bibliographystyle{plainnat}
\bibliography{main}

\clearpage
\beginappendix
\section{Notation}
\label{app:notation}

Table~\ref{tab:notation} collects only the recurring method symbols. Symbols used in one proof are defined where that proof begins.

\begin{table}[h]
\caption{Core symbols used by the method.}
\label{tab:notation}
\begin{center}
\small
\fitwidth{%
\begin{tabular}{ll}
\toprule
Symbol & Meaning \\
\midrule
$t,s,\tau$ & input, policy-evaluation and generic proof times, with noise at $1$ and data at $0$ \\
$\vx_0,\veps,\vx_t$ & data, Gaussian noise and their interpolation in \eqref{eq:path} \\
$\hat\vx_0,\hat\vx_t$ & generated endpoint and its noised state \\
$c,s_{\mathrm{cfg}}$ & condition and classifier-free guidance scale \\
$\vv_T,\vv_{T,\mathrm{cfg}}$ & teacher velocity and its guided version \\
$\theta,\bar\theta$ & student parameters and their moving average \\
$\gP_\theta,\{\vv_k,\pi_k,\gamma_k\}_{k=1}^{K}$ & policy and its $K$ momentum modes \\
$G_\theta,\tilde\vv_\theta$ & analytic endpoint generator and momentum-mixture velocity parameterization \\
$q_{\theta,t},p_{T,t},\vv_{q_\theta,t},\vv_{p_T,t}$ & noised laws and their marginal velocities \\
$\lambda_{\mathrm{tr}},\lambda_{\mathrm{dm}},\lambda_{\mathrm{fm}},\lambda_{\mathrm{dp}}$ & fixed loss weights in \eqref{eq:total} \\
$\rho,\lambda(\cdot),w_{\mathrm D}(\cdot)$ & time sampling and time-dependent distribution weights \\
$a_i,r,\rho_r$ & few-step stage times, sampled rollout length and stage-conditioned time distribution \\
$\mathcal S_t,\hat\vx_{t,s}^{T},\vv_{t,s}^{\theta},\vv_{t,s}^{T}$ & trajectory grid, teacher states and student/teacher grid velocities \\
$\sg[\cdot]$ & stop-gradient \\
\bottomrule
\end{tabular}
}
\end{center}
\end{table}

\section{Figure and table conventions}
\label{app:conventions}

\paragraph{Overview figure.}
In the left panel of \figref{fig:overview}, the labeled arrows at the noised student state are the start-time velocities $\vv_{t,t}^{\theta}$ and $\vv_{t,t}^{T}$ whose difference forms the distribution matching direction of \eqref{eq:ldm}. The vertical double arrows mark the trajectory grid points at which \eqref{eq:ltr} compares $\vv_{t,s}^{\theta}$ with $\vv_{t,s}^{T}$ for $s\in\mathcal S_t$, and the two dashed paths end at the student and teacher endpoint distributions. In the right panel, the time axis is logarithmic, all methods share the training settings of \secref{sec:setup}, and line styles and markers identify the methods in grayscale.

\paragraph{Metric abbreviations and table styling.}
Recall abbreviates Recall-DINOv2, Diversity and DINO Div.\ abbreviate DINO Diversity, and Pick, HPS, ImgRwd and Aesth abbreviate PickScore, HPSv2, ImageReward and the aesthetic score. The extra-model and auxiliary-network columns count trainable parameter sets besides the student and exclude the frozen teacher. In the gray teacher reference rows, NFE is written as sampling steps times the two classifier-free-guidance branches, and FD-DINOv2 and Recall-DINOv2 are omitted because the teacher images define their reference. Within each NFE block of Tables~\ref{tab:sana} and~\ref{tab:sd35}, the best and second-best student values are bolded and underlined; the teacher row does not take part in this ranking. The last column of Table~\ref{tab:cost} reports the normalized time needed to reach the FD-DINOv2 target level of the right panel of \figref{fig:overview}. The upper block of Table~\ref{tab:policy} varies $K$ at the default $L$, its lower block varies $L$ at the default $K$, and its parameter column reports additional trainable parameters as a percentage of the teacher parameter count.

\paragraph{Mechanism figure.}
The left panel of \figref{fig:mechanism} compares students distilled from Stable Diffusion 3.5 Medium at checkpoints of equal training length. Each method generates one sample for each of 512 held-out test prompts with a single unguided student step, and both methods share the prompts and the noise drawn for \eqref{eq:noised}, so the two curves are paired. At each noising time, the start-time estimate $\vv_{t,t}^{\theta}$ is compared by cosine distance with the sample-conditioned velocity $\veps-\hat\vx_0$, where lower values indicate stronger directional agreement. The curves cross between $t=0.24$ and $t=0.22$, after which TRD has the lower distance toward $t=0$. The time grid holds 51 points spaced by $0.02$ from $t=1$ to $t=0$. Each curve is the median over samples, each band spans the 25th to the 75th percentile, and line styles and markers identify the methods in grayscale. The right panel compares matched one-step TRD variants with and without trajectory alignment at the same training checkpoint. For each of 512 held-out prompts, both students and the teacher start from the same noise; the analytic student paths and the teacher path are compared at the same 51 times. The teacher path uses 100 Euler evaluations with a guidance scale of $4.5$, while each student uses one unguided evaluation. Each curve is the mean latent RMSE over prompts, and each band is a pointwise 95\% percentile bootstrap interval over prompts from 2{,}000 resamples.

\paragraph{Training-loss figure.}
\figref{fig:sd3-losses} reports every loss after applying its objective weight. All curves are solid, and their markers identify the four objectives in grayscale.

\paragraph{Policy-head distributions.}
\figref{fig:momentum} uses a $K=4$ Stable Diffusion 3.5 Medium checkpoint without trajectory alignment. We evaluate one unguided conditional forward at $t=1$ for each of 512 held-out prompts, using one initial-noise sample per prompt. Gate weights and momentum rates use separate pooled vertical ranges. Every head within one panel shares the same range, and no samples are clipped or rescaled by head. Violin width represents kernel density, the thin and thick vertical segments span the 5th to 95th and 25th to 75th percentiles, and the white marker denotes the median. All gate distributions remain well above zero and the rate distributions remain separated, so neither plotted output exhibits collapse at $t=1$. These statistics are descriptive and do not identify training-induced specialization or per-head agreement with the teacher trajectory.

\paragraph{Qualitative figures.}
Figures~\ref{fig:qualitative-sana} and~\ref{fig:qualitative-sd35} use prompts from the held-out test split of the publicly available dataset \citep{schuhmann2022laion}, and samples for the same prompt and sampling budget share their initial noise across methods. Figures~\ref{fig:additional-sana} and~\ref{fig:additional-sd35} show additional one-step TRD samples for the two teachers.

\section{Technical analysis and proofs}
\label{app:proofs}

This appendix proves the statements collected in Proposition~\ref{prop:ideal} and then describes what remains once the ideal assumptions are removed. The argument runs in three stages that depend on each other in this order. The first stage concerns the policy alone: one forward pass returns coefficients that define an explicit function of time, so the induced trajectory is available in closed form, and Proposition~\ref{prop:grid} records how many independent velocities such a field can carry on a finite grid. The second stage identifies what the two distribution terms optimize: Proposition~\ref{prop:selffm} turns the shared student output into the marginal velocity of the current student law and converts velocities into scores, Theorem~\ref{thm:dmd} uses both facts to show that the detached inner product of \eqref{eq:ldm} differentiates to a time-averaged reverse Kullback-Leibler gradient, Proposition~\ref{prop:compromise} shows that the trajectory target on the same start-time output only rescales that gradient, and Theorem~\ref{thm:endpoint} shows that the zeros of that divergence are exactly the students whose endpoint law is the teacher endpoint law. The third stage replaces exact quantities by errors: Theorem~\ref{thm:policy} bounds the endpoint error of the analytic rollout by the grid loss that is actually minimized, Corollary~\ref{cor:joint} adds the trajectory error and the distribution error in a single Wasserstein bound, and Proposition~\ref{prop:dp} states what distribution preservation controls. Appendix~\ref{app:qualifications} then removes the remaining idealizations, namely the unguided teacher and exact optimization.

\paragraph{Conventions.}
All statements hold for almost every condition $c$, which is suppressed unless it is averaged explicitly. The generated endpoint $\hat\vx_0=G_\theta(\vx_1,c)$ and its noised state $\hat\vx_t$ from \eqref{eq:noised} are treated as random variables with laws $q_{\theta,0}$ and $q_{\theta,t}$, and the arguments of velocity and score fields are dropped whenever those fields are evaluated at the sampled state. For an endpoint law $\chi$ and $t\in(0,1)$, the Gaussian noising operator, the marginal velocity and the score are
\begin{equation}
\begin{aligned}
\vX&\sim\chi,\quad \veps\sim\gN(\vzero,\mI)\ \text{independently},\quad
\vX_t=(1-t)\vX+t\veps,\quad \mathcal N_t\chi=\Law(\vX_t),\\
\vv_{\chi,t}(\vx)&=\E\big[\veps-\vX\mid\vX_t=\vx\big],
\qquad
\vs_{\chi,t}(\vx)=\nabla_{\vx}\log\big(\mathcal N_t\chi\big)(\vx).
\end{aligned}
\label{eq:noisingop}
\end{equation}
Applying \eqref{eq:noisingop} to $\chi=q_{\theta,0}$ and $\chi=p_{T,0}$ gives the noised laws $q_{\theta,t}$ and $p_{T,t}$ and the four fields $\vv_{q_\theta,t}$, $\vv_{p_T,t}$, $\vs_{q_\theta,t}$ and $\vs_{p_T,t}$ used below. We write $J_\theta G_\theta$ for the Jacobian of the generator in $\theta$ and $\|\cdot\|_{\op}$ for the operator norm.

\begin{assumption}[Regularity]
\label{ass:reg}
The following hold for almost every condition $c$. (i) $\rho$ is supported in $[t_{\min},t_{\max}]\subset(0,1)$. (ii) $\theta\mapsto G_\theta(\vx_1,c)$ is differentiable with $\E\|J_\theta G_\theta\|_{\op}^2<\infty$ and $\E\|G_\theta\|_2^2<\infty$. (iii) For every $t\in(0,1)$ the laws $q_{\theta,t}$ and $p_{T,t}$ have positive densities that are continuously differentiable in $\vx$, and $\KL(q_{\theta,t}\Vert p_{T,t})<\infty$. (iv) Differentiation in $\theta$ may be exchanged with the integrals below, and the boundary terms of the integrations by parts vanish at infinity. (v) $\lambda(\cdot)$ and $t\lambda(t)/(1-t)$ are non-negative and $\rho$-integrable.
\end{assumption}

Item (i) keeps the velocity-to-score conversion away from its singularity at $t=0$ and keeps the noised law informative about the endpoint at $t=1$; items (ii) to (iv) are what the gradient computation in Theorem~\ref{thm:dmd} uses; item (v) makes the weighted divergence in \eqref{eq:jd} finite.

\subsection{The policy field and its reach on a trajectory grid}
\label{app:policyfield}

The coefficients returned by one forward pass are held fixed over the interval, so the velocity in \eqref{eq:policy} is an explicit function of time and can be integrated in closed form. With
\begin{equation}
C(\gamma,\Delta):=\int_0^{\Delta}\gamma^{u}\,\mathrm{d}u=
\begin{cases}
(\gamma^{\Delta}-1)/\log\gamma, & \gamma\neq1,\\
\Delta, & \gamma=1,
\end{cases}
\label{eq:kernel}
\end{equation}
integrating \eqref{eq:policy} backward from $\vx(t)=\vx_t$ gives the state at any earlier time,
\begin{equation}
\vx(\tau)=\vx_t-\sum_{k=1}^{K}\pi_k\vv_k\gamma_k^{\,1-t}
C(\gamma_k,t-\tau),\qquad \tau\le t,
\label{eq:rollout}
\end{equation}
and the endpoint generator of \secref{sec:preliminaries} is $G_\theta(\vx_t,t,c)=\vx(0)$. \citet{yang2026arcflow} derive \eqref{eq:rollout}, its constant-velocity limit at $\gamma_k=1$ and exact finite-grid interpolation with sufficiently many modes. TRD adopts the rollout and finite-grid supervision. We record the full rank statement to cover both undercomplete and overcomplete policies; TRD's methodological change is to couple this policy with distribution matching and self velocity estimation.

Fix a start $(\vx_t,t,c)$ and write the distinct grid times as $\mathcal S_t=\{\tau_0,\ldots,\tau_M\}$, where $t=\tau_0>\cdots>\tau_M\ge0$. Collect the mode coefficients at the start into $\vm_k=\pi_k\vv_k\gamma_k^{1-t}$ and define the evaluation matrix $\mV\in\R^{(M+1)\times K}$ with entries $\mV_{jk}=\gamma_k^{\,t-\tau_j}$, so that the grid velocities of \eqref{eq:ltr} are the linear images
\begin{equation}
\vv_{t,\tau_j}^{\theta}=\tilde\vv_\theta(\tau_j;\gP_\theta(\vx_t,t,c))=\sum_{k=1}^{K}\mV_{jk}\vm_k,
\qquad j=0,\dots,M .
\label{eq:gridlinear}
\end{equation}

\begin{proposition}[Grid rank of the ArcFlow policy]
\label{prop:grid}
If the grid times are distinct and $\gamma_1,\dots,\gamma_K$ are positive and pairwise distinct, then $\rank\mV=\min\{M+1,K\}$. In particular, $K\ge M+1$ makes \eqref{eq:gridlinear} solvable for any prescribed grid velocities, and for an equally spaced grid of spacing $h_{\mathrm{grid}}$ with $K=M+1$ the matrix $\mV$ is a Vandermonde matrix in the values $\gamma_k^{h_{\mathrm{grid}}}$, with $\det\mV=\pm\prod_{p<q}(\gamma_q^{h_{\mathrm{grid}}}-\gamma_p^{h_{\mathrm{grid}}})\neq0$.
\end{proposition}

\begin{proof}
Write $\mV_{jk}=e^{\kappa_ku_j}$ with $\kappa_k=\log\gamma_k$ and $u_j=t-\tau_j$, so that both families are pairwise distinct real numbers. The proof uses one elementary fact: a function $u\mapsto\sum_{k=1}^{K}c_ke^{\kappa_ku}$ with distinct real $\kappa_k$ and some $c_k\ne0$ has at most $K-1$ real zeros. This follows by induction on $K$, because dividing by $e^{\kappa_1u}$ leaves the zeros unchanged and differentiating then removes the first term while, by Rolle's theorem, losing at most one zero.

If $K\le M+1$ and $\mV\vc=\vzero$ for some $\vc\ne\vzero$, the function $u\mapsto\sum_kc_ke^{\kappa_ku}$ vanishes at the $M+1\ge K$ distinct points $u_j$, contradicting that fact, so the columns are independent and $\rank\mV=K$. If $K>M+1$ and $\vb^{\top}\mV=\vzero^{\top}$ for some $\vb\ne\vzero$, the function $\kappa\mapsto\sum_jb_je^{\kappa u_j}$ has $M+1$ terms with distinct rates $u_j$ and vanishes at the $K>M+1$ distinct points $\kappa_k$, again a contradiction, so the rows are independent and $\rank\mV=M+1$. Full row rank makes \eqref{eq:gridlinear} solvable in $\{\vm_k\}$ coordinate by coordinate, and any solution is realizable by the policy, because the softmax keeps $\pi_k>0$ while the base velocities $\vv_k$ are unconstrained. On an equally spaced grid $u_j=jh_{\mathrm{grid}}$, so $\mV_{jk}=(\gamma_k^{h_{\mathrm{grid}}})^{j}$ and the determinant is the Vandermonde determinant.
\end{proof}

The interpolation consequence of Proposition~\ref{prop:grid} recovers the ArcFlow finite-grid result, while the full rank formula also states the undercomplete case. It is an existence statement at one fixed start and does not assert that training finds the interpolating coefficients. The rank formula also identifies conditioning limits: when $K\le M+1$, two momentum rates approaching each other drive the smallest singular value toward zero; when $M+1\le K$, two grid times approaching each other have the analogous effect. The coefficients that realize prescribed grid velocities can therefore become sensitive to supervision error. The result supports the single-forward design only when the relevant rates and grid times remain separated.

\subsection{What the distribution terms optimize}
\label{app:proofdist}

\paragraph{Self flow matching and the velocity-to-score dictionary.}
The first step fixes the meaning of the shared student output. Self flow matching regresses $\vv_{t,t}^{\theta}$ on the conditional velocity $\veps-\hat\vx_0$ of the generated sample, and the following statement identifies the resulting population target and records the identity that converts velocities into scores on the path \eqref{eq:path}.

\begin{proposition}[Self flow matching identifies the student marginal velocity]
\label{prop:selffm}
For every square-integrable velocity predictor $\vv$,
\begin{equation}
\E\big\|\vv(\hat\vx_t,t,c)-(\veps-\hat\vx_0)\big\|_2^2
=\E\big\|\vv(\hat\vx_t,t,c)-\vv_{q_\theta,t}\big\|_2^2
+\E\big\|(\veps-\hat\vx_0)-\vv_{q_\theta,t}\big\|_2^2 ,
\label{eq:fmdecomp}
\end{equation}
so $\vv_{q_\theta,t}$ is the almost everywhere unique minimizer and the excess risk is exactly the squared velocity estimation error. Moreover, for endpoint laws $\chi$ and $\zeta$ noised by \eqref{eq:noisingop} and every $t\in(0,1)$,
\begin{equation}
\vs_{\chi,t}(\vx)=-\frac{\vx+(1-t)\vv_{\chi,t}(\vx)}{t},
\qquad
\vs_{\chi,t}-\vs_{\zeta,t}=-\frac{1-t}{t}\big(\vv_{\chi,t}-\vv_{\zeta,t}\big).
\label{eq:v2s}
\end{equation}
\end{proposition}

\begin{proof}
Both claims follow from conditioning on the noised state. For \eqref{eq:fmdecomp}, add and subtract $\vv_{q_\theta,t}(\hat\vx_t)$ inside the norm and expand the square. By \eqref{eq:noisingop} that field is the conditional mean of $\veps-\hat\vx_0$ given $\hat\vx_t$, so the residual $(\veps-\hat\vx_0)-\vv_{q_\theta,t}$ is orthogonal to every square-integrable function of $(\hat\vx_t,t,c)$ and the cross term vanishes. Only the first term depends on $\vv$, and it is zero exactly when $\vv=\vv_{q_\theta,t}$ up to null sets.

For \eqref{eq:v2s}, differentiate the Gaussian mixture density of $\vX_t$ under the integral sign:
\begin{equation}
\vs_{\chi,t}(\vx)
=\E\left[-\frac{\vx-(1-t)\vX}{t^2}\,\bigg|\,\vX_t=\vx\right]
=-\frac1t\,\E\big[\veps\mid\vX_t=\vx\big],
\label{eq:scoreid}
\end{equation}
where the second equality uses $\vx-(1-t)\vX=t\veps$ on the event $\vX_t=\vx$. Since $\veps=\vX_t+(1-t)(\veps-\vX)$, the same conditioning gives $\E[\veps\mid\vX_t=\vx]=\vx+(1-t)\vv_{\chi,t}(\vx)$, which is the first identity in \eqref{eq:v2s}. Writing that identity for $\chi$ and for $\zeta$ and subtracting removes the common term $-\vx/t$ and leaves the second.
\end{proof}

Equation~\ref{eq:fmdecomp} is what allows the student to act as its own velocity estimator: the loss that trains the estimator has a single population target, and the gap that remains when the target is not reached is exactly the quantity that Theorem~\ref{thm:dmd} charges to the distribution gradient. Equation~\ref{eq:v2s} is the dictionary between the two descriptions of a noised law, and the factor $(1-t)/t$ that it produces is the part of $\lambda(t)$ absorbed in the main text. Both are statements about functions, and whether the shared network realizes them remains an assumption.

\begin{remark}[Why the self flow matching inputs are detached]
\label{rem:sg}
At the optimal predictor, the value of \eqref{eq:lfm} is the second term of \eqref{eq:fmdecomp}, the conditional variance of the regression target given the noised state. Since $\veps-\hat\vx_0=(\hat\vx_t-\hat\vx_0)/t$, that value equals $t^{-2}\E\|\hat\vx_0-\E[\hat\vx_0\mid\hat\vx_t]\|_2^2$, and for $t\in(0,1)$ the Gaussian likelihood is positive everywhere, so the conditional law of $\hat\vx_0$ spreads over the whole support of $q_{\theta,0}$. The residual is therefore zero only when $q_{\theta,0}$ is a point mass. If gradients were allowed to flow through the generated sample, the generator would have a direct route to lower this loss by collapsing, so \eqref{eq:lfm} detaches both the input state and the regression target.
\end{remark}

\paragraph{The distribution matching update is a weighted reverse Kullback-Leibler gradient.}
The second step connects the detached update of \eqref{eq:ldm} to the divergence between the noised laws in \eqref{eq:jd}. Its time weight and the effective velocity weight satisfy
\begin{equation}
w_{\mathrm D}(t)=\frac{t\lambda(t)}{1-t}.
\label{eq:weightrelation}
\end{equation}
Let $\ve_\theta=\vv_{t,t}^{\theta}-\vv_{q_\theta,t}$ and $\ve_T=\vv_{t,t}^{T}-\vv_{p_T,t}$ be the student and teacher velocity errors at the sampled state, with second moments bounded by $\delta_\theta^2$ and $\delta_T^2$.

\begin{theorem}[The detached update is a weighted reverse Kullback-Leibler gradient]
\label{thm:dmd}
When $\ve_\theta=\ve_T=\vzero$, the automatic-differentiation gradient of \eqref{eq:ldm} equals the gradient of \eqref{eq:jd}, and
\begin{equation}
\nabla_\theta\gJ_{\mathrm D}(\theta)
=-\E\Big[\lambda(t)(1-t)\big(J_\theta G_\theta\big)^{\!\top}
\big(\vv_{q_\theta,t}-\vv_{p_T,t}\big)\Big].
\label{eq:dmdgrad}
\end{equation}
In general the expected update $\widehat\vg$ obtained from estimated velocities obeys
\begin{equation}
\big\|\widehat\vg-\nabla_\theta\gJ_{\mathrm D}(\theta)\big\|_2
\le C_v(\delta_\theta+\delta_T),
\qquad
C_v^2=\E\big[\lambda(t)^2(1-t)^2
\|J_\theta G_\theta\|_{\op}^2\big].
\label{eq:bias}
\end{equation}
\end{theorem}

\begin{proof}
Fix $(t,c)$ and let $\vb_t(\vx)=\E[J_\theta\hat\vx_t\mid\hat\vx_t=\vx]$ be the velocity that a parameter perturbation induces on the sample. Then
\begin{align}
\nabla_\theta\KL(q_{\theta,t}\Vert p_{T,t})
&=\int\big(\nabla_\theta q_{\theta,t}\big)
\log\frac{q_{\theta,t}}{p_{T,t}}\,\mathrm d\vx
\label{eq:klstep1}\\
&=-\int\nabla_{\vx}\!\cdot\!\big(q_{\theta,t}\vb_t\big)
\log\frac{q_{\theta,t}}{p_{T,t}}\,\mathrm d\vx
\label{eq:klstep2}\\
&=\int q_{\theta,t}\,\vb_t^{\top}
\big(\vs_{q_\theta,t}-\vs_{p_T,t}\big)\,\mathrm d\vx
\label{eq:klstep3}\\
&=\E\Big[\big(J_\theta\hat\vx_t\big)^{\!\top}
\big(\vs_{q_\theta,t}-\vs_{p_T,t}\big)\Big]
\label{eq:klstep4}\\
&=-\frac{1-t}{t}\,
\E\Big[\big(J_\theta\hat\vx_t\big)^{\!\top}
\big(\vv_{q_\theta,t}-\vv_{p_T,t}\big)\Big].
\label{eq:klstep5}
\end{align}
Weighting \eqref{eq:klstep5} by $w_{\mathrm D}(t)$ from \eqref{eq:weightrelation}, averaging over $(t,c)$ and substituting $J_\theta\hat\vx_t=(1-t)J_\theta G_\theta$ gives \eqref{eq:dmdgrad}. Automatic differentiation of \eqref{eq:ldm} contracts the same Jacobian with the detached vector $-\lambda(t)(\vv_{t,t}^{\theta}-\vv_{t,t}^{T})$, so
\begin{equation}
\widehat\vg
=-\E\big[\lambda(t)(1-t)(J_\theta G_\theta)^{\top}(\vv_{t,t}^{\theta}-\vv_{t,t}^{T})\big]
=\nabla_\theta\gJ_{\mathrm D}(\theta)
-\E\big[\lambda(t)(1-t)(J_\theta G_\theta)^{\top}(\ve_\theta-\ve_T)\big].
\label{eq:adgrad}
\end{equation}

Step \eqref{eq:klstep1} differentiates under the integral sign and drops the term $\int\nabla_\theta q_{\theta,t}\,\mathrm d\vx=\nabla_\theta1=0$ produced by the derivative of the logarithm. Step \eqref{eq:klstep2} is the continuity equation, which states that the law of $\hat\vx_t$ moves under the parameter perturbation by the transport of $q_{\theta,t}$ along $\vb_t$. Step \eqref{eq:klstep3} integrates by parts, uses $\nabla_{\vx}\log(q_{\theta,t}/p_{T,t})=\vs_{q_\theta,t}-\vs_{p_T,t}$ and discards the boundary term by Assumption~\ref{ass:reg}(iv). Step \eqref{eq:klstep4} rewrites the integral as an expectation and replaces $\vb_t$ by $J_\theta\hat\vx_t$ under the tower property, which is valid because the other factor depends on the sample only through $\hat\vx_t$. Step \eqref{eq:klstep5} applies the score difference in \eqref{eq:v2s}, and the weight relation \eqref{eq:weightrelation} is exactly the factor that turns $(1-t)^2/t$ into the $\lambda(t)(1-t)$ of \eqref{eq:dmdgrad}. The first claim is the case $\ve_\theta=\ve_T=\vzero$ of \eqref{eq:adgrad}. For \eqref{eq:bias}, take norms in the error term of \eqref{eq:adgrad}, bound $\|(J_\theta G_\theta)^{\top}\vu\|_2\le\|J_\theta G_\theta\|_{\op}\|\vu\|_2$, apply Cauchy-Schwarz to the expectation, and use $(\E\|\ve_\theta-\ve_T\|_2^2)^{1/2}\le\delta_\theta+\delta_T$.
\end{proof}

Two properties of \eqref{eq:bias} matter in practice. The bias is linear in the velocity errors, so the shared estimator trained by \eqref{eq:lfm} controls it directly through \eqref{eq:fmdecomp}, and no separate network is needed to make the statement meaningful. The constant $C_v$ carries $\lambda(t)$, which contains the score-to-velocity factor $(1-t)/t$ of \eqref{eq:v2s} and grows without bound as $t\to0$, which is why $\rho$ is truncated away from the data endpoint in Assumption~\ref{ass:reg}(i). Appendix~\ref{app:qualifications} turns this bias into an optimization statement.

\paragraph{Trajectory alignment at the start time rescales the update.}
The grid $\mathcal S_t$ contains $t$, so the start-time output $\vv_{t,t}^{\theta}$ receives the teacher target $\vv_{t,t}^{T}$ from \eqref{eq:ltr} in addition to the regression target of \eqref{eq:lfm}, and a nonzero trajectory weight makes $\ve_\theta$ in Theorem~\ref{thm:dmd} nonzero even at the population optimum. The next statement shows that this error has a special form. Fix $(t,c)$ and write $\vu(\vx)$ for the start-time output evaluated at the detached state $\vx=\sg[\hat\vx_t]$, so that $\vu(\hat\vx_t)=\vv_{t,t}^{\theta}$. Let $\vv_T(\cdot,t,c)$ be the teacher field that supplies $\vv_{t,t}^{T}$ in both \eqref{eq:ltr} and \eqref{eq:ldm}, and let $b=\lambda_{\mathrm{tr}}/(M+1)$ be the weight that \eqref{eq:total} places on the start-time term of \eqref{eq:ltr}, where $M$ may depend on $t$.

\begin{proposition}[The start-time compromise preserves the distribution direction]
\label{prop:compromise}
Assume $\lambda_{\mathrm{fm}}>0$ and that $\vu$ ranges over square-integrable functions of $\hat\vx_t$ independently of the other grid velocities and of the generator. The start-time terms of $\lambda_{\mathrm{fm}}\gL_{\mathrm{fm}}+\lambda_{\mathrm{tr}}\gL_{\mathrm{tr}}$ are minimized, uniquely up to null sets, by
\begin{equation}
\vu^\star=\frac{\lambda_{\mathrm{fm}}\vv_{q_\theta,t}+b\,\vv_T}{\lambda_{\mathrm{fm}}+b},
\qquad
\vu^\star-\vv_T=\alpha\big(\vv_{q_\theta,t}-\vv_T\big),
\qquad
\alpha=\frac{\lambda_{\mathrm{fm}}}{\lambda_{\mathrm{fm}}+b}\in\Big[\frac{\lambda_{\mathrm{fm}}}{\lambda_{\mathrm{fm}}+\lambda_{\mathrm{tr}}},1\Big].
\label{eq:compromise}
\end{equation}
If in addition $\vv_T=\vv_{p_T,t}$ and the student reports $\vv_{t,t}^{\theta}=\vu^\star(\hat\vx_t)$, the automatic-differentiation gradient of \eqref{eq:ldm} equals the gradient of \eqref{eq:jd} with $w_{\mathrm D}(t)$ replaced by $\alpha\,w_{\mathrm D}(t)$.
\end{proposition}

\begin{proof}
Write $a=\lambda_{\mathrm{fm}}$ and abbreviate the per-sample start-time objective and its population value as
\begin{align}
&\E\Big[a\big\|\vu-(\veps-\hat\vx_0)\big\|_2^2+b\big\|\vu-\vv_T\big\|_2^2\Big]
\label{eq:cmp1}\\
&=\E\Big[a\big\|\vu-\vv_{q_\theta,t}\big\|_2^2+b\big\|\vu-\vv_T\big\|_2^2\Big]
+a\,\E\big\|(\veps-\hat\vx_0)-\vv_{q_\theta,t}\big\|_2^2
\label{eq:cmp2}\\
&=(a+b)\,\E\big\|\vu-\vu^\star\big\|_2^2
+\frac{ab}{a+b}\,\E\big\|\vv_{q_\theta,t}-\vv_T\big\|_2^2
+a\,\E\big\|(\veps-\hat\vx_0)-\vv_{q_\theta,t}\big\|_2^2 .
\label{eq:cmp3}
\end{align}
Step \eqref{eq:cmp1} collects the start-time term of \eqref{eq:lfm} and the $s=t$ summand of \eqref{eq:ltr} with their weights from \eqref{eq:total}. Step \eqref{eq:cmp2} applies the decomposition \eqref{eq:fmdecomp} to the first term; the second term is unchanged because $\vv_T$ is a function of $\hat\vx_t$. Step \eqref{eq:cmp3} completes the square pointwise through $a\|\vu-\vp\|_2^2+b\|\vu-\vq\|_2^2=(a+b)\|\vu-(a\vp+b\vq)/(a+b)\|_2^2+ab\|\vp-\vq\|_2^2/(a+b)$ with $\vp=\vv_{q_\theta,t}$ and $\vq=\vv_T$. Only the first term of \eqref{eq:cmp3} depends on $\vu$, which gives the unique minimizer and, after subtracting $\vv_T$, the second identity in \eqref{eq:compromise}. The range of $\alpha$ follows from $0<b\le\lambda_{\mathrm{tr}}$. For the gradient claim, substituting $\vu^\star-\vv_{p_T,t}=\alpha(\vv_{q_\theta,t}-\vv_{p_T,t})$ into the detached vector of \eqref{eq:ldm} turns it into the exact vector of Theorem~\ref{thm:dmd} with $\lambda(t)$ replaced by $\alpha\lambda(t)$, and \eqref{eq:weightrelation} carries the factor $\alpha$ to $w_{\mathrm D}(t)$.
\end{proof}

Proposition~\ref{prop:compromise} controls the part of $\ve_\theta$ that trajectory alignment introduces: at the joint optimum this part equals $-(1-\alpha)(\vv_{q_\theta,t}-\vv_{p_T,t})$, which is parallel to the exact distribution direction, so it rescales the time weight of \eqref{eq:jd} without changing the zeros used in Theorem~\ref{thm:endpoint} or the ideal equilibrium of Proposition~\ref{prop:ideal}. The middle term of \eqref{eq:cmp3} also shows that the two start-time residuals cannot vanish together before $\vv_{q_\theta,t}=\vv_T$. The statement relies on the start-time output being free. With shared parameters, and in particular when $K<M+1$ couples the start-time value to the other grid velocities through \eqref{eq:gridlinear}, the realized error also contains a component that is not parallel to the distribution direction, and Theorem~\ref{thm:dmd} bounds its effect.

\paragraph{A zero of the divergence forces endpoint agreement.}
Theorem~\ref{thm:dmd} concerns noised laws at times inside $(0,1)$, while the quantity of interest is the generated endpoint law. The next statement closes that gap at a global optimum.

\begin{theorem}[Vanishing divergence forces endpoint agreement]
\label{thm:endpoint}
Assume $p_{T,t}=\mathcal N_tp_{T,0}$, some $\theta^\dagger$ satisfies $q_{\theta^\dagger,0}=p_{T,0}$ for almost every $c$, and $\rho(\{t<1:\lambda(t)>0\})>0$. Then $\inf_\theta\gJ_{\mathrm D}(\theta)=0$, and every global minimizer $\theta^{\mathrm{opt}}$ satisfies $q_{\theta^{\mathrm{opt}},0}=p_{T,0}$ for almost every $c$.
\end{theorem}

\begin{proof}
The realizing parameter $\theta^\dagger$ makes every noisy divergence vanish, so the infimum is zero. Since the integrand of \eqref{eq:jd} is non-negative, a global minimizer satisfies $q_{\theta^{\mathrm{opt}},t}=p_{T,t}$ for $\rho$-almost every $t<1$ with $\lambda(t)>0$, and by Fubini's theorem such a $t$ exists for almost every $c$. Fix one. Both laws are the same Gaussian convolution of their endpoint laws, so their characteristic functions factor as
\begin{equation}
\varphi_{\mathcal N_t\chi}(\vxi)
=\varphi_{\chi}\big((1-t)\vxi\big)
\exp\!\left(-\tfrac12t^2\|\vxi\|_2^2\right).
\label{eq:cf}
\end{equation}
The Gaussian factor never vanishes and $1-t>0$, so the two endpoint characteristic functions agree on all of $\R^d$, and the uniqueness theorem for characteristic functions gives the claim.
\end{proof}

Two hypotheses carry the conclusion. Both noisy families must come from noising one endpoint law, which is the hypothesis that classifier-free guidance breaks and that Appendix~\ref{app:qualifications} examines, and the positively weighted times must include some $t<1$, since the noised law at $t=1$ is pure noise and retains no endpoint information. The conclusion also concerns the exact global minimum only. Noising contracts distances between endpoint laws, so a small positive value of $\gJ_{\mathrm D}$ does not by itself bound the endpoint discrepancy, and the statement that survives at finite error is Corollary~\ref{cor:joint} rather than a perturbed version of this theorem.

\subsection{Finite-error behaviour and the complementarity of the two blocks}
\label{app:prooffinite}

\paragraph{From grid velocity error to endpoint error.}
Theorem~\ref{thm:dmd} already shows what inexact velocities cost the distribution gradient. The corresponding question for trajectory alignment is what the grid loss of \eqref{eq:ltr}, which compares velocities at $M+1$ times only, implies about the distance between the student and teacher paths at all times.

Fix a realization of the training start $(\hat\vx_t,t,c)$ and hold the predicted policy $\gP_\theta(\sg[\hat\vx_t],t,c)$ fixed over the interval. Let $\vx_\theta(\tau)$ be the analytic student rollout \eqref{eq:rollout} from that start and let $\vx_T(\tau)$ be the exact teacher solution from the same start, so that both satisfy $\vx_\theta(t)=\vx_T(t)=\hat\vx_t$. Write
\[
\vv_{T,\mathrm{path}}(\tau)=\vv_T(\vx_T(\tau),\tau,c),
\qquad
\ve(\tau)=\tilde\vv_\theta(\tau;\gP_\theta(\sg[\hat\vx_t],t,c))-\vv_{T,\mathrm{path}}(\tau)
\]
for the teacher velocity along its own path and for the velocity discrepancy at time $\tau$, and note that $\ve(\tau_j)=\vv_{t,\tau_j}^{\theta}-\vv_{T,\mathrm{path}}(\tau_j)$ differs from the residual seen by training only through the accuracy of the teacher targets. For one sample, let $\ell_{\mathrm{tr}}=(M+1)^{-1}\sum_{j=0}^{M}\|\vv_{t,\tau_j}^{\theta}-\vv_{t,\tau_j}^{T}\|_2^2$ be the bracketed quantity of \eqref{eq:ltr}, let $\{A_j\}_{j=0}^{M}$ be a partition of $[0,t]$ with $\tau_j\in A_j$, and set $\Delta_{\max}=\max_j|A_j|$ and $\Delta_{\mathrm{mesh}}=\max_j\sup_{\tau\in A_j}|\tau-\tau_j|$.

\begin{theorem}[Endpoint error of the single-forward rollout]
\label{thm:policy}
If $\ve\in L^2([0,t];\R^d)$, then
\begin{equation}
\vx_\theta(\tau)-\vx_T(\tau)=-\int_\tau^{t}\ve(u)\,\mathrm{d}u,
\qquad
\big\|\vx_\theta(0)-\vx_T(0)\big\|_2
\le\sqrt{t}\,\big\|\ve\big\|_{L^2([0,t])}.
\label{eq:pathbound}
\end{equation}
If in addition $\ve$ is $L_e$-Lipschitz in time and the teacher targets satisfy $\max_j\|\vv_{t,\tau_j}^{T}-\vv_{T,\mathrm{path}}(\tau_j)\|_2\le\eta_{\max}$, then
\begin{equation}
\big\|\vx_\theta(0)-\vx_T(0)\big\|_2
\le
\sqrt{t\,\Delta_{\max}(M+1)\ell_{\mathrm{tr}}}
+t\big(\eta_{\max}+L_e\Delta_{\mathrm{mesh}}\big),
\label{eq:gridbound}
\end{equation}
whose right-hand side becomes $t(\sqrt{\ell_{\mathrm{tr}}}+\eta_{\max}+L_e\Delta_{\mathrm{mesh}})$ when the cells have equal length. If the teacher targets come from an order $p_{\mathrm{sol}}$ solver with maximum step $h_{\mathrm{sol}}$ and global error constant $C_{\mathrm{sol}}$, and the teacher field is $L_x$-Lipschitz in the state, then $\eta_{\max}\le L_xC_{\mathrm{sol}}h_{\mathrm{sol}}^{p_{\mathrm{sol}}}$.
\end{theorem}

\begin{proof}
The two rollouts share their state at time $t$, so subtracting their integral forms gives the identity in \eqref{eq:pathbound}, and Cauchy-Schwarz applied to $\int_0^t\|\ve(u)\|_2\,\mathrm du$ gives the inequality. No Lipschitz condition on the teacher field and no Gr\"onwall factor appear here, because $\ve$ compares the student policy with the teacher velocity along the teacher path rather than at the student state.

For the finite grid, control $\ve$ on each cell by its value at the grid point it contains:
\begin{align}
\sup_{\tau\in A_j}\|\ve(\tau)\|_2
&\le\|\vv_{t,\tau_j}^{\theta}-\vv_{t,\tau_j}^{T}\|_2+\eta_{\max}+L_e\Delta_{\mathrm{mesh}},
\label{eq:cellbound}\\
\|\ve\|_{L^2([0,t])}
&\le\Big(\sum_{j=0}^{M}|A_j|\,\|\vv_{t,\tau_j}^{\theta}-\vv_{t,\tau_j}^{T}\|_2^2\Big)^{1/2}
+\sqrt t\,\big(\eta_{\max}+L_e\Delta_{\mathrm{mesh}}\big)
\label{eq:l2step1}\\
&\le\sqrt{\Delta_{\max}(M+1)\ell_{\mathrm{tr}}}
+\sqrt t\,\big(\eta_{\max}+L_e\Delta_{\mathrm{mesh}}\big).
\label{eq:l2step2}
\end{align}
Step \eqref{eq:cellbound} compares $\ve(\tau)$ with $\ve(\tau_j)$ through the Lipschitz property over a distance of at most $\Delta_{\mathrm{mesh}}$, then inserts $\vv_{t,\tau_j}^{T}$ into $\ve(\tau_j)=\vv_{t,\tau_j}^{\theta}-\vv_{T,\mathrm{path}}(\tau_j)$ and applies the target error bound $\eta_{\max}$. Step \eqref{eq:l2step1} integrates the square of \eqref{eq:cellbound} over each cell and applies Minkowski's inequality in the sequence space weighted by the cell lengths, where the constant contributes $(\sum_j|A_j|)^{1/2}=\sqrt t$. Step \eqref{eq:l2step2} replaces every $|A_j|$ by $\Delta_{\max}$ and uses $\sum_j\|\vv_{t,\tau_j}^{\theta}-\vv_{t,\tau_j}^{T}\|_2^2=(M+1)\ell_{\mathrm{tr}}$. Combining \eqref{eq:l2step2} with \eqref{eq:pathbound} gives \eqref{eq:gridbound}, and equal cells have $\Delta_{\max}=t/(M+1)$. For the last claim, the solver state error is at most $C_{\mathrm{sol}}h_{\mathrm{sol}}^{p_{\mathrm{sol}}}$ at every grid point, and evaluating the $L_x$-Lipschitz teacher field at those states turns it into a velocity error of at most $L_xC_{\mathrm{sol}}h_{\mathrm{sol}}^{p_{\mathrm{sol}}}$.
\end{proof}

Equation~\ref{eq:gridbound} separates three error sources that training treats very differently. The first term is the grid residual, which is what \eqref{eq:ltr} minimizes. The second is the accuracy of the teacher targets, which is set by the solver and its step size rather than by the student. The third is the variation of the discrepancy between grid points, which no discrete loss can observe: without a regularity condition in time, a bump placed between two grid points leaves $\ell_{\mathrm{tr}}$ unchanged and moves the endpoint arbitrarily far. The policy of \eqref{eq:policy} is a sum of $K$ smooth exponentials, which is the structural reason a modest grid is informative about the whole interval, and Proposition~\ref{prop:grid} is the complementary statement that such a field can still match all grid points.

The configurations in Table~\ref{tab:hparams} cap the grid span, so a single grid may end at $s_M>0$. The identity in \eqref{eq:pathbound} holds at every $\tau\le t$, and repeating the argument on a partition of $[s_M,t]$ gives the bounds of Theorem~\ref{thm:policy} for the state at $s_M$ with $t$ replaced by $t-s_M$. On $[0,s_M]$ the grid of that sample provides no supervision, and the full-interval bound for that sample relies on $L_e$ over the long final cell. Lower times are supervised in other iterations, for policies started from states with smaller $t\sim\rho$.

\paragraph{Joint control of the law produced by the student policy.}
The two supervision blocks act on different objects: trajectory alignment compares maps from a common start, while distribution matching compares laws at a common time. The next statement puts both errors in one bound. Let $\Psi_\theta^{t}$ be the analytic policy map from time $t$ to time $0$, and let $\Phi_T^{t}$ be the exact teacher flow map over the same interval.

\begin{corollary}[Joint bound on the law produced by the student policy]
\label{cor:joint}
Assume $(\Phi_T^{t})_{\#}p_{T,t}=p_{T,0}$, $\Lip(\Phi_T^{t})\le L_\Phi$, and finite second moments. Define the mean squared trajectory map error from the current student law,
\begin{equation}
\varepsilon_{\mathrm{tr}}^2(t)
=\E_{\hat\vx_t\sim q_{\theta,t}}
\big\|\Psi_\theta^{t}(\hat\vx_t)-\Phi_T^{t}(\hat\vx_t)\big\|_2^2 .
\label{eq:trajmaperror}
\end{equation}
Then
\begin{equation}
W_2\big((\Psi_\theta^{t})_{\#}q_{\theta,t},p_{T,0}\big)
\le\varepsilon_{\mathrm{tr}}(t)+L_\Phi W_2(q_{\theta,t},p_{T,t}),
\label{eq:jointw2}
\end{equation}
and if $p_{T,t}$ satisfies a Talagrand $T_2(C_T)$ inequality \citep{otto2000generalization}, then
\begin{equation}
W_2\big((\Psi_\theta^{t})_{\#}q_{\theta,t},p_{T,0}\big)
\le\varepsilon_{\mathrm{tr}}(t)
+L_\Phi\sqrt{2C_T\KL(q_{\theta,t}\Vert p_{T,t})}.
\label{eq:jointkl}
\end{equation}
\end{corollary}

\begin{proof}
Both terms come from the triangle inequality for $W_2$ applied at the intermediate law $(\Phi_T^{t})_{\#}q_{\theta,t}$. For the first term, couple $(\Psi_\theta^{t})_{\#}q_{\theta,t}$ and $(\Phi_T^{t})_{\#}q_{\theta,t}$ by sending the same $\hat\vx_t\sim q_{\theta,t}$ through both maps; the cost of that particular coupling is $\varepsilon_{\mathrm{tr}}^2(t)$, and $W_2$ is the infimum over all couplings. For the second term, push an optimal coupling of $q_{\theta,t}$ and $p_{T,t}$ through $\Phi_T^{t}$; the map is $L_\Phi$-Lipschitz, so the cost grows by at most $L_\Phi$, and the assumption $(\Phi_T^{t})_{\#}p_{T,t}=p_{T,0}$ identifies the second marginal. Equation~\ref{eq:jointkl} then substitutes the Talagrand inequality $W_2^2(q_{\theta,t},p_{T,t})\le2C_T\KL(q_{\theta,t}\Vert p_{T,t})$.
\end{proof}

The two summands of \eqref{eq:jointw2} are exactly the objects that the two training blocks control, and neither controls the other. Trajectory alignment bounds the first through \eqref{eq:gridbound}, which applies to the integrand of \eqref{eq:trajmaperror} sample by sample, since $\Psi_\theta^t$ is the analytic rollout and $\Phi_T^t$ the exact teacher solution from the same start. Distribution matching bounds the second through the divergence of \eqref{eq:jd}, in the form \eqref{eq:jointkl}. A student whose noised law already matches the teacher law still needs an accurate map to turn it into samples, and an accurate map applied to a mismatched law reproduces the mismatch, which is the technical content of combining the two signals. The bound also states its own limit: the controlled object is the law obtained by pushing $q_{\theta,t}$ through the student policy, which need not equal the generator endpoint law $q_{\theta,0}$, so this is a complementarity statement rather than endpoint recovery at finite error.

\paragraph{What distribution preservation controls.}
The remaining term compares the current generator with its moving-average reference at a fixed $\bar\theta$. Since it is a squared distance between two maps evaluated at the same latent input, it is a proximal penalty in the space of generator maps, and it dominates the corresponding distances between the generated laws.

\begin{proposition}[What distribution preservation controls]
\label{prop:dp}
Let $\gR_{\mathrm{dp}}(\theta;\bar\theta)=\E_{\vx_1,c}\|G_\theta(\vx_1,c)-G_{\bar\theta}(\vx_1,c)\|_2^2$, which equals \eqref{eq:ldp} when $\bar\theta$ is held fixed. Then
\begin{equation}
\E_c\big[W_2^2(q_{\theta,0}(\cdot\mid c),q_{\bar\theta,0}(\cdot\mid c))\big]\le\gR_{\mathrm{dp}},
\qquad
\E_c\big[W_2^2(q_{\theta,t}(\cdot\mid c),q_{\bar\theta,t}(\cdot\mid c))\big]\le(1-t)^2\gR_{\mathrm{dp}},
\label{eq:dpw2}
\end{equation}
and for every $t\in(0,1)$,
\begin{equation}
\E_c\Big[\max\big\{\KL(q_{\theta,t}(\cdot\mid c)\Vert q_{\bar\theta,t}(\cdot\mid c)),\,\KL(q_{\bar\theta,t}(\cdot\mid c)\Vert q_{\theta,t}(\cdot\mid c))\big\}\Big]\le\frac{(1-t)^2}{2t^2}\gR_{\mathrm{dp}}.
\label{eq:dpkl}
\end{equation}
\end{proposition}

\begin{proof}
Fix $c$. Sending the same latent $\vx_1$ through both generators is one coupling of $q_{\theta,0}(\cdot\mid c)$ and $q_{\bar\theta,0}(\cdot\mid c)$, and its cost is the summand of $\gR_{\mathrm{dp}}$, which gives the first bound in \eqref{eq:dpw2}. Sharing the Gaussian noise of \eqref{eq:noised} as well makes the two noised samples differ by $(1-t)(G_\theta-G_{\bar\theta})$, which gives the second bound.

For \eqref{eq:dpkl}, keep the shared latent in the description and compare the joint laws of $(\vx_1,\hat\vx_t)$ under $\theta$ and under $\bar\theta$. The latent has the same law in both, so the KL between the joint laws is the average of the conditional KL given $\vx_1$. Conditioned on the latent, the two noised outputs are Gaussians with the same covariance $t^2\mI$ and means differing by $(1-t)(G_\theta-G_{\bar\theta})$, so their KL in either direction is the squared mean difference divided by $2t^2$. Discarding the latent cannot increase the KL, which transfers the bound to the marginal laws of $\hat\vx_t$, and averaging over $c$ finishes the proof.
\end{proof}

Proposition~\ref{prop:dp} is written for the one-step coupling in
\eqref{eq:ldp}. For Algorithm~\ref{alg:fewstep}, the same argument applies
conditionally on the shared final-stage input
$(\vx_{a_{r-1}},a_{r-1},c)$ after replacing the student and moving-average
generators by $G_\theta(\vx_{a_{r-1}},a_{r-1},c)$ and
$G_{\bar\theta}(\vx_{a_{r-1}},a_{r-1},c)$, respectively.

Both bounds can be strict, because $\gR_{\mathrm{dp}}$ fixes one latent coupling: two generators that induce the same law but permute which latent produces which sample have zero distance between their laws and a positive preservation loss. This is intended, since the term is meant to limit how far one update moves the generator map, and not to measure a distance between distributions. The divergence bound in \eqref{eq:dpkl} also degrades as $t\to0$, so it is informative only at the noise levels that $\rho$ actually visits.

\section{Guidance and optimization qualifications}
\label{app:qualifications}

Appendix~\ref{app:proofs} assumes an unguided teacher and, in Proposition~\ref{prop:ideal}, exact optimization of an expressive enough student. This appendix removes those assumptions one at a time. Classifier-free guidance changes the object that the distribution term matches, the residual velocity error of Theorem~\ref{thm:dmd} sets a floor on the stationarity that term can reach, and the four updates together with the moving average form a coupled dynamical system whose equilibria have to be examined separately.

\paragraph{Classifier-free guidance.}
Classifier-free guidance \citep{ho2022classifier} replaces the teacher field in \eqref{eq:teacherroll} and \eqref{eq:ldm} by
\begin{equation}
\vv_{T,\mathrm{cfg}}(\vx,t,c)
=\vv_T(\vx,t,\varnothing)
+s_{\mathrm{cfg}}\big[\vv_T(\vx,t,c)-\vv_T(\vx,t,\varnothing)\big].
\label{eq:cfg}
\end{equation}
Let $p_{T,\mathrm{c},t}(\vx\mid c)$ and $p_{T,\mathrm{u},t}(\vx)$ be the conditional and unconditional teacher marginals, with scores $\vs_{T,\mathrm{c},t}$ and $\vs_{T,\mathrm{u},t}$. If both branches predict exact marginal velocities under the same noising kernel, \eqref{eq:v2s} gives
\begin{equation}
\vs_{T,\mathrm{cfg},t}(\vx,c)
=(1-s_{\mathrm{cfg}})\vs_{T,\mathrm{u},t}(\vx)
+s_{\mathrm{cfg}}\vs_{T,\mathrm{c},t}(\vx\mid c)
=-\frac{\vx+(1-t)\vv_{T,\mathrm{cfg}}(\vx,t,c)}{t}.
\label{eq:cfgscore}
\end{equation}
When normalizable, \eqref{eq:cfgscore} is the score of the geometric
combination $p_{T,\mathrm{c},t}^{\,s_{\mathrm{cfg}}}
p_{T,\mathrm{u},t}^{\,1-s_{\mathrm{cfg}}}$, and Theorem~\ref{thm:dmd}
therefore applies separately at each time, with $p_{T,t}$ read as that
combination. For guidance scales outside $[0,1]$, normalizability
additionally requires suitable tail control.

What guidance breaks is the hypothesis of Theorem~\ref{thm:endpoint}. These
time-indexed geometric combinations are in general not the noised versions of
one endpoint law, so guided TRD matches a family of per-time score fields
that need not describe a single target distribution. The endpoint conclusion
applies directly to the conditional branch at $s_{\mathrm{cfg}}=1$ and to the
unconditional branch at $s_{\mathrm{cfg}}=0$, and other scales require
cross-time consistency as an additional assumption. The trajectory results of
Appendix~\ref{app:prooffinite} are unaffected, because they treat the guided
field as the field being integrated. Guidance evaluates both teacher
branches, which gives $2(M+1)$ teacher evaluations per iteration.

\paragraph{Stationarity floor from velocity estimation error.}
Theorem~\ref{thm:dmd} bounds the bias of the distribution update by
$\beta_{\mathrm{vel}}=C_v(\delta_\theta+\delta_T)$. The standard descent argument for biased
stochastic gradients turns that bias into a floor on the stationarity that
this term can reach.

\begin{proposition}[Stationarity floor under a biased update]
\label{prop:biasfloor}
Assume $\gJ_{\mathrm D}$ is bounded below by $\gJ_{\inf}$ and has an
$L_{\mathrm D}$-Lipschitz gradient, the stochastic update $\widehat\vg_k$ has conditional
mean $\nabla\gJ_{\mathrm D}(\theta_k)+\vdelta_k$ with
$\|\vdelta_k\|_2\le\beta_{\mathrm{vel}}$ and conditional variance at most $\sigma_g^2$, and
the step size satisfies $0<\eta_{\mathrm{opt}}\le1/(2L_{\mathrm D})$. Then the iterates
$\theta_{k+1}=\theta_k-\eta_{\mathrm{opt}}\widehat\vg_k$ obey
\begin{equation}
\frac1N\sum_{k=0}^{N-1}\E\big\|\nabla\gJ_{\mathrm D}(\theta_k)\big\|_2^2
\le\frac{4\big(\gJ_{\mathrm D}(\theta_0)-\gJ_{\inf}\big)}{\eta_{\mathrm{opt}} N}
+6\beta_{\mathrm{vel}}^2+2L_{\mathrm D}\eta_{\mathrm{opt}}\sigma_g^2 .
\label{eq:biasfloor}
\end{equation}
\end{proposition}

\begin{proof}
Write $\vg_k=\nabla\gJ_{\mathrm D}(\theta_k)$. The descent lemma for an
$L_{\mathrm D}$-smooth function gives
$\E_k[\gJ_{\mathrm D}(\theta_{k+1})]\le\gJ_{\mathrm D}(\theta_k)
-\eta_{\mathrm{opt}}\vg_k^{\top}(\vg_k+\vdelta_k)+\tfrac12L_{\mathrm D}\eta_{\mathrm{opt}}^2\E_k\|\widehat\vg_k\|_2^2$.
Bound the cross term by
$-\vg_k^{\top}\vdelta_k\le\tfrac14\|\vg_k\|_2^2+\|\vdelta_k\|_2^2$ and the
second moment by
$\E_k\|\widehat\vg_k\|_2^2\le2\|\vg_k\|_2^2+2\|\vdelta_k\|_2^2+\sigma_g^2$.
With $L_{\mathrm D}\eta_{\mathrm{opt}}\le1/2$, the coefficient of $\|\vg_k\|_2^2$ is at most $-\eta_{\mathrm{opt}}/4$
and the coefficient of $\|\vdelta_k\|_2^2$ at most $3\eta_{\mathrm{opt}}/2$, so
$\tfrac{\eta_{\mathrm{opt}}}{4}\E\|\vg_k\|_2^2\le\E[\gJ_{\mathrm D}(\theta_k)
-\gJ_{\mathrm D}(\theta_{k+1})]+\tfrac32\eta_{\mathrm{opt}}\beta_{\mathrm{vel}}^2
+\tfrac12L_{\mathrm D}\eta_{\mathrm{opt}}^2\sigma_g^2$. Summing over $k$, telescoping and using the
lower bound on $\gJ_{\mathrm D}$ gives \eqref{eq:biasfloor}.
\end{proof}

Only the first term of \eqref{eq:biasfloor} decreases with $N$, so the bound
retains the non-vanishing term $6\beta_{\mathrm{vel}}^2$ under persistent velocity error and
does not establish convergence of the average gradient to zero. The constant
$C_v$ in \eqref{eq:bias} weights the velocity error by the time weight and the
generator Jacobian, so the time range and weighting determine how strongly
estimation error affects the update. The statement concerns the distribution
term in isolation and does not imply convergence of the full TRD dynamics.

\paragraph{Equilibria and local stability of the full dynamics.}
Let $\vh(\theta)$ denote the expected weighted update from self flow matching,
distribution matching and trajectory alignment under the stop-gradient rules
of Algorithm~\ref{alg:trd}, and let $\vd(\theta,\bar\theta)$ denote the
expected preservation update, with the weights of \eqref{eq:total} absorbed
into both. These are vector fields rather than gradients of one fixed scalar
objective, because the states at which they are evaluated are produced by the
current student and then detached. The associated mean dynamics are
\begin{equation}
\dot\theta=-\vh(\theta)-\vd(\theta,\bar\theta),
\qquad
\dot{\bar\theta}=r_{\mathrm{ema}}\big(\theta-\bar\theta\big),
\qquad r_{\mathrm{ema}}>0 ,
\label{eq:meandyn}
\end{equation}
where $r_{\mathrm{ema}}$ is the positive continuous-time rate corresponding
to the discrete decay $\mu$; a unit-step forward-Euler correspondence gives
$r_{\mathrm{ema}}=1-\mu$.

\begin{proposition}[Equilibria and limits of the complete mean field]
\label{prop:fulllimit}
Assume $\vh$ and $\vd$ are continuous and $\vd(\theta,\theta)=\vzero$. Every
convergent solution of \eqref{eq:meandyn} has a limit with
$\bar\theta_\infty=\theta_\infty$ and $\vh(\theta_\infty)=\vzero$. In the
ideal unguided setting, any $\theta^\star$ that simultaneously attains
$q_{\theta^\star,0}=p_{T,0}$, its own marginal velocity and zero trajectory
residual gives an equilibrium $(\theta^\star,\theta^\star)$. A general zero of
$\vh$ need not have these three properties, because the weighted updates can
cancel each other.
\end{proposition}

\begin{proof}
At the ideal target, Proposition~\ref{prop:selffm} makes the expected self
flow matching update zero, Theorem~\ref{thm:dmd} makes the distribution update
zero because the two velocity fields coincide, and the trajectory residual is
zero by assumption; preservation vanishes because $\bar\theta=\theta$. This
proves the equilibrium claim. For a convergent solution of a continuous
autonomous system, the vector field vanishes at the limit. The second equation
of \eqref{eq:meandyn} gives $\bar\theta_\infty=\theta_\infty$, and
substituting this into the first equation together with
$\vd(\theta,\theta)=\vzero$ gives $\vh(\theta_\infty)=\vzero$.
\end{proof}

\begin{theorem}[Local stability of the full dynamics at the ideal target]
\label{thm:stab}
Let $\theta^\star$ be an ideal equilibrium as in
Proposition~\ref{prop:fulllimit}. Assume $\vh$ and $\vd$ are continuously
differentiable nearby, that the symmetric part of
$\mA=\nabla\vh(\theta^\star)$ is bounded below by $m\mI$ for some $m>0$, and
that the preservation field linearizes as $\mB(\theta-\bar\theta)$ with
$\mB\succeq0$. Then $(\theta^\star,\theta^\star)$ is locally asymptotically
stable, so every mean trajectory started close enough to it converges to it,
and along that limit the student endpoint law equals $p_{T,0}$, its velocity
estimate is exact and its trajectory residual vanishes.
\end{theorem}

\begin{proof}
Let $\delta\theta=\theta-\theta^\star$ and $\delta\bar\theta=\bar\theta-\theta^\star$. Linearizing \eqref{eq:meandyn} at the equilibrium gives the system matrix
\begin{equation}
\mL=\begin{bmatrix}-(\mA+\mB) & \mB\\ r_{\mathrm{ema}}\mI & -r_{\mathrm{ema}}\mI\end{bmatrix},
\label{eq:lin}
\end{equation}
so it suffices to show that every eigenvalue of $\mL$ has negative real part. Suppose instead that $\mL\,(\vx,\vy)^{\top}=z\,(\vx,\vy)^{\top}$ for some eigenvector and some $z$ with $\operatorname{Re}(z)\ge0$. Then $z\ne-r_{\mathrm{ema}}$, so the second block row gives $\vy=r_{\mathrm{ema}}\vx/(z+r_{\mathrm{ema}})$, and $\vx\ne\vzero$, since $\vx=\vzero$ would force $\vy=\vzero$. Substituting $\vy$ into the first block row gives
\[
\left[z\mI+\mA+\frac{z}{z+r_{\mathrm{ema}}}\mB\right]\vx=\vzero .
\]
Take the inner product with $\vx$ and keep the real part. The first term contributes $\operatorname{Re}(z)\|\vx\|_2^2\ge0$. The second contributes at least $m\|\vx\|_2^2>0$, because only the symmetric part of $\mA$ survives in the real part. The third contributes $\operatorname{Re}\big(z/(z+r_{\mathrm{ema}})\big)\,\vx^{*}\mB\vx\ge0$, because $\mB\succeq0$ and because $\operatorname{Re}(z)\ge0$ with $r_{\mathrm{ema}}>0$ makes that scalar non-negative. The real part is therefore strictly positive, which contradicts the displayed equation. Hence $\mL$ is Hurwitz and the Lyapunov indirect method gives local asymptotic stability. For \eqref{eq:ldp}, the preservation field linearizes to $\mB=2\lambda_{\mathrm{dp}}\E[(J_\theta G_\theta)^{\top}J_\theta G_\theta]$ at $\theta=\bar\theta$, which is positive semidefinite as required.
\end{proof}

The conclusion is local, and the stability margin has to come from $\vh$: the
preservation term contributes a positive semidefinite $\mB$, which can
preserve an existing margin and cannot create one. The ordinary differential
equation method relates decreasing-step stochastic updates to the mean
dynamics under bounded iterates, Robbins and Monro step sizes, martingale
difference noise and step-size-weighted summable bias
\citep{borkar2008stochastic}. Convergence to this particular equilibrium also
requires eventual entry into its attraction basin. Persistent estimation error
or a fixed step size prevents direct application of that result, and
\eqref{eq:biasfloor} only quantifies stationarity of the distribution term in
isolation. Taken together with Corollary~\ref{cor:joint}, the supported
qualifications are that the ideal target is locally attracting under an
explicit margin condition and that finite error yields a joint bound rather
than endpoint recovery from a small noisy divergence.

\section{Implementation details}
\label{app:impl}

This appendix records the TRD training and evaluation settings in Table~\ref{tab:hparams}. SANA-1.6B is a linear diffusion transformer with a deeply compressed autoencoder and a decoder-only text encoder \citep{xie2025sana}. Stable Diffusion 3.5 Medium follows the multimodal rectified-flow transformer family introduced by \citet{esser2024scaling}; Table~\ref{tab:hparams} gives the exact checkpoint identifier without assigning a parameter count whose published sources use different scopes. For both teachers, the last $L$ student blocks are duplicated across the $K$ policy modes of \eqref{eq:policy}, and only these policy heads and the backbone adapters are trained. The baseline rows use the adaptations described in \secref{sec:setup}; they are not reproductions of the original papers' training regimes. Unless an ablation states otherwise, the listed settings are fixed.

\paragraph{Prompt split and evaluation protocol.}
All image experiments use prompts from a publicly available dataset \citep{schuhmann2022laion}. We partition the prompts into disjoint training and test sets, reserve 2{,}500 prompts for testing and use the remainder for training. Every reported metric uses this test set. Each method generates four images per prompt under a fixed four-seed schedule shared across methods, yielding 10{,}000 images, and the corresponding teacher generates a matched 10{,}000-image reference set.

\paragraph{Metrics.}
We extract DINOv2 ViT-L/14 features with registers \citep{oquab2024dinov2,darcet2024vision} from the method and teacher images. FD-DINOv2 fits a Gaussian to each feature set and computes their Fr\'echet distance, following the DINOv2 evaluation of \citet{stein2023exposing} and the original Fr\'echet metric \citep{heusel2017gans}. Recall-DINOv2 applies the improved precision and recall estimator to measure coverage of the teacher feature manifold \citep{kynkaanniemi2019improved}. DINO Diversity adapts the mean pairwise DINOv2 cosine distance of \citet{sorokin2025imagerefl} to four images per prompt. We estimate human preference with PickScore \citep{kirstain2023pick}, HPSv2 \citep{wu2023human} and ImageReward \citep{xu2023imagereward}, visual appeal with LAION-Aesthetics Predictor V2 on CLIP ViT-L/14 features \citep{schuhmann2022laion,radford2021learning}, and text alignment with $100$ times the non-negative cosine similarity of OpenAI CLIP ViT-B/32 embeddings \citep{radford2021learning}. Table~\ref{tab:hparams} records the specified model variants and DINO preprocessing.

\paragraph{Few-step training.}
Algorithm~\ref{alg:fewstep} states the few-step iteration of \secref{sec:algo}, whose outer noising interval follows the sampled step count. For a two-step schedule $1=a_0>a_1>a_2=0$, one-step rollouts draw the outer noising time from $[a_1,a_0]$, while two-step rollouts draw it from $[a_2,a_1]$.

\begin{algorithm}[h]
\caption{One TRD training iteration for few-step distillation with random-length consistency sampling}
\label{alg:fewstep}
\begin{algorithmic}[1]
\REQUIRE student $\theta$, moving average $\bar\theta$ with rate $\mu$, frozen teacher $\vv_T$, consistency-sampling times $1=a_0>\cdots>a_R=0$, stage-conditioned time distributions $\{\rho_r\}_{r=1}^{R}$ with $\operatorname{supp}(\rho_r)\subseteq[a_r,a_{r-1}]$, grid $\mathcal S_t=\{t,s_1,\ldots,s_M\}$ with $t>s_1>\cdots>s_M\ge0$, loss weights $\lambda_{\mathrm{tr}},\lambda_{\mathrm{dm}},\lambda_{\mathrm{fm}},\lambda_{\mathrm{dp}}$ of \eqref{eq:total} and time weight $\lambda(t)$
\STATE sample $c$, initial noise $\vx_{a_0}\sim\gN(\vzero,\mI)$ and step count $r\sim\operatorname{Unif}\{1,\dots,R\}$
\FOR{$i=0,\dots,r-2$}
    \STATE {\bfseries advance the sampler, without gradients:} predict $\hat\vx_0^{(i)}=G_\theta(\vx_{a_i},a_i,c)$, draw $\veps_{i+1}\sim\gN(\vzero,\mI)$ and re-noise it to $\vx_{a_{i+1}}=(1-a_{i+1})\hat\vx_0^{(i)}+a_{i+1}\veps_{i+1}$
\ENDFOR
\STATE {\bfseries generate:} evaluate the student at $(\vx_{a_{r-1}},a_{r-1},c)$ and integrate the output velocity by \eqref{eq:rollout} to $\hat\vx_0=G_\theta(\vx_{a_{r-1}},a_{r-1},c)$, the only endpoint prediction that retains gradients
\STATE {\bfseries re-noise:} draw $t\sim\rho_r$ and independent $\veps\sim\gN(\vzero,\mI)$; form $\hat\vx_t=(1-t)\hat\vx_0+t\veps$, retaining its gradient path to $\theta$
\STATE {\bfseries predict velocities:} evaluate the student at $(\sg[\hat\vx_t],t,c)$; obtain $\{\vv_{t,s}^{\theta}:s\in\mathcal S_t\}$ from \eqref{eq:policy}, without further student evaluations
\STATE {\bfseries teacher rollout, without gradients:} initialize $\hat\vx_{t,t}^{T}=\hat\vx_t$; for decreasing $s\in\mathcal S_t$, evaluate $\vv_{t,s}^{T}$ and, if a next grid time $s'$ exists, advance to $\hat\vx_{t,s'}^{T}$ by \eqref{eq:teacherroll}
\STATE {\bfseries reference endpoint, without gradients:} evaluate $G_{\bar\theta}(\vx_{a_{r-1}},a_{r-1},c)$ on the same final-stage input
\STATE {\bfseries form the losses:} use the grid velocities for \eqref{eq:ltr}, $\vv_{t,t}^{\theta}$ and $\veps-\sg[\hat\vx_0]$ for \eqref{eq:lfm}, the detached start-time velocity difference and $\hat\vx_t$ for \eqref{eq:ldm}, and the paired final-stage endpoints for \eqref{eq:ldp}
\STATE {\bfseries update:} backpropagate \eqref{eq:total}, take one optimizer step on $\theta$, then set $\bar\theta\leftarrow\mu\bar\theta+(1-\mu)\theta$
\end{algorithmic}
\end{algorithm}

{\footnotesize
\setlength{\tabcolsep}{3pt}
\begin{longtable}{@{}>{\raggedright\arraybackslash}p{0.28\linewidth}>{\raggedright\arraybackslash}p{0.33\linewidth}>{\raggedright\arraybackslash}p{0.33\linewidth}@{}}
\caption{Configuration of the TRD image experiments.}
\label{tab:hparams}\\
\toprule
Item & SANA-1.6B & SD3.5 Medium \\
\midrule
\endfirsthead
\toprule
Item & SANA-1.6B & SD3.5 Medium \\
\midrule
\endhead
\bottomrule
\endfoot
\bottomrule
\endlastfoot

\multicolumn{3}{@{}l}{\textit{Teacher and reference generation}}\\*
Checkpoint & \path{Efficient-Large-Model/Sana_1600M_1024px_BF16_diffusers} & \path{stabilityai/stable-diffusion-3.5-medium} \\
Reference scheduler and steps & checkpoint scheduler; $20$ steps & FlowMatch Euler scheduler; $40$ steps \\
Reference guidance scale & $4.5$ & $4.5$ \\
Conditioning and image resolution & frozen Gemma2 text encoder; $1024\times1024$ & three frozen checkpoint text encoders; $1024\times1024$ \\
\midrule

\multicolumn{3}{@{}l}{\textit{Student policy}}\\*
Number of modes $K$ & $4$ & $4$ \\
Duplicated final blocks $L$ and output branches & $L=1$; velocity, gate and momentum projections per mode & $L=1$; velocity, gate and momentum projections per mode \\
LoRA rank, $\alpha$ and dropout & $64$; $64$; $0$ & $32$; $32$; $0$ \\
LoRA targets and trainable components & all backbone linear layers except the output projection; full policy heads & attention projections in all transformer blocks; full policy heads \\
Teacher and student parameter sharing & one pretrained backbone; student adapters are disabled for teacher queries & one pretrained backbone; student adapters are disabled for teacher queries \\
Policy initialization & copied final block and teacher output projection; zero gate logits; $\gamma_k$ log-spaced in $[0.4,5.0]$ with one head at $1$ & copied final block and teacher output projection; zero gate logits; $\gamma_k$ log-spaced in $[0.8,1.2]$ with one head at $1$ \\
\midrule

\multicolumn{3}{@{}l}{\textit{TRD objective}}\\*
Loss weights $(\lambda_{\mathrm{tr}},\lambda_{\mathrm{dm}},\lambda_{\mathrm{fm}},\lambda_{\mathrm{dp}})$ & $(0.1,10.0,1.0,1.0)$ & $(0.1,30.0,1.0,1.0)$ \\
Effective time weight $\lambda(t)$ & constant $1$ with mean reduction & constant $1$ with mean reduction \\
Preservation reference and coupling & refreshed every step with EMA rate $\mu=0.9$; shared final-stage input & refreshed every step with EMA rate $\mu=0.9$; shared final-stage input \\
Teacher and student guidance scales & $4.5$; $1.0$ & $4.5$; $1.0$ \\
Noising-time distribution & $\operatorname{Unif}(0,1)$ & uniform within the selected rollout interval, with raw-time clip $[10,980]$ \\
Trajectory grid and solver & explicit Euler; step $0.05$; maximum span $0.30$; at most $7$ grid points & explicit Euler; step $0.025$; maximum span $0.20$; at most $9$ grid points \\
Few-step consistency schedule & scheduler-generated times; rollout length sampled uniformly from $\{1,\ldots,R\}$ & $R=2$ with raw times $(1000,780)$; rollout depth induced by the clipped interval widths \\
Momentum scale and $|\log\gamma_k|$ clamp & $1.0$; $20.0$ & $1.0$; $20.0$ \\
\midrule

\multicolumn{3}{@{}l}{\textit{Optimization}}\\*
Learning rates (backbone/basic; gate; momentum) & $10^{-4}$; $10^{-5}$; $10^{-5}$ & $10^{-4}$; $10^{-5}$; $10^{-5}$ \\
Optimizer and schedule & 8-bit AdamW \citep{loshchilov2019decoupled,dettmers2022optimizers}; $(\beta_1,\beta_2)=(0.9,0.999)$; weight decay $10^{-4}$; constant; no warmup & same \\
Batch, precision and clipping & global batch $8$; bfloat16; gradient norm $1.0$ & global batch $8$; bfloat16; gradient norm $1.0$ \\
\midrule

\multicolumn{3}{@{}l}{\textit{Evaluation}}\\*
Held-out prompts and seeds & $2{,}500$ disjoint prompts selected with dataset seed $666$; image seeds start at $42$ & same \\
Sample count & four images per prompt and $10{,}000$ student images; matched teacher reference set & same \\
FD-DINOv2 features and preprocessing & \path{dinov2_vitl14_reg}; resize to $256$, center crop to $224$, ImageNet normalization & same \\
Recall-DINOv2 & student $3$-NN manifold balls; coverage of the matched $10{,}000$-image teacher set & same \\
DINO Diversity & DINOv2 features; mean pairwise cosine distance within each four-image prompt group & same \\
Preference and aesthetic models & PickScore v1; HPSv2; ImageReward v1.0; LAION-Aesthetics Predictor V2 with CLIP ViT-L/14 \citep{schuhmann2022laion,radford2021learning} & same \\
CLIP similarity backbone & OpenAI CLIP ViT-B/32; $100\max(\operatorname{cos},0)$ & same \\
\end{longtable}
}

\section{Per-iteration cost}
\label{app:cost}

This appendix accounts for the network evaluations and the persistent state of one TRD iteration, and compares them with a distribution distillation method that trains a separate fake score model. The counts follow Algorithm~\ref{alg:trd} under the settings of Table~\ref{tab:hparams}; \secref{sec:cost} reports what they cost in measured memory and time.

\paragraph{Network evaluations.}
One iteration evaluates the student twice, once at $(\vx_1,1,c)$ for generation and once at the detached noised state $(\sg[\hat\vx_t],t,c)$ for the shared velocity predictions, and backpropagates through both. The frozen teacher is evaluated $M+1$ times, once at each grid point of the rollout in \eqref{eq:teacherroll}, and the moving-average copy once at $(\vx_1,1,c)$. Only the two student evaluations carry gradients, so the teacher rollout and the moving-average pass need no stored activations. The grid size therefore enters the cost through frozen forward passes alone. Because $\gP_\theta(\sg[\hat\vx_t],t,c)$ gives $\vv_{t,s}^{\theta}$ for every $s\in\mathcal S_t$ in closed form, supervising $M+1$ grid points costs no additional student evaluation, whereas supervising the same points with a student that predicts one velocity per query would require one student pass per point. With classifier-free guidance each grid point evaluates the conditional and the unconditional branch, which raises the teacher count to $2(M+1)$ as recorded in Appendix~\ref{app:qualifications}.

\paragraph{Parameters and optimizer state.}
TRD trains the policy heads and the low-rank backbone adapters and keeps the pretrained backbone frozen, so gradients and optimizer state exist only for that trainable subset. Teacher queries reuse the same backbone with the student adapters disabled, and no second copy of the backbone weights is held for the teacher. The moving-average copy tracks the trainable subset as well, so distribution preservation adds one parameter copy of that size and one forward pass, without gradients or optimizer state.

\paragraph{Comparison with a trained fake score model.}
A method that estimates the student score with a separate network holds that network at full size together with its gradients and optimizer state and updates it with its own forward and backward passes. The DMD2 measurement uses five fake-score updates per generator update, matching its ImageNet and SDXL configurations under the two-time-scale rule of \citet{yin2024improved}; this ratio is a configuration choice rather than a universal property of DMD2. Estimating the same score with self flow matching on the shared policy removes those updates and that state, and adds in their place the $M+1$ frozen teacher passes of the trajectory rollout and the single moving-average pass. Table~\ref{tab:cost} and the right panel of \figref{fig:overview} report where this exchange leaves the measured memory, per-iteration time and time to a target FD-DINOv2 level.

\paragraph{Few-step variant.}
Algorithm~\ref{alg:fewstep} prepends $r-1$ gradient-free student passes to the same iteration, where $r$ is drawn uniformly from $\{1,\dots,R\}$, so the expected additional cost is $(R-1)/2$ student forward passes per iteration. All remaining counts are unchanged after these detached passes. In particular, the moving average still requires one forward pass at $(\vx_{a_{r-1}},a_{r-1},c)$; for $r=1$, this reduces to the one-step input $(\vx_1,1,c)$.

\section{Additional results}
\label{app:results}

\begin{figure}[t]
\centering
\includegraphics[width=\linewidth]{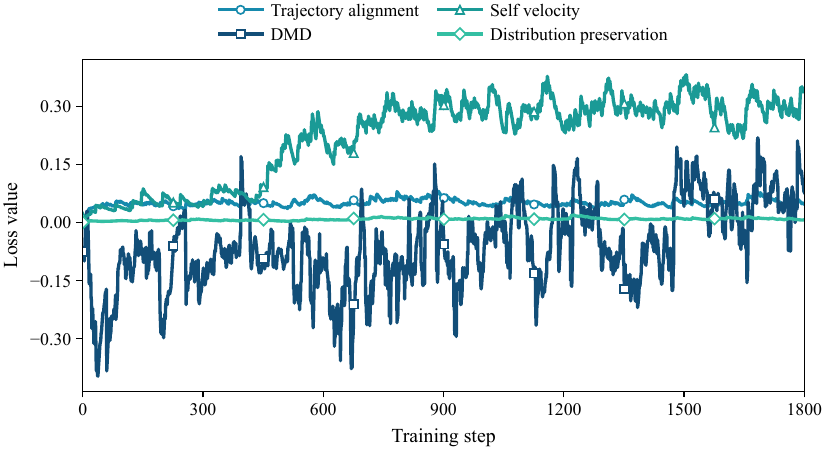}
\caption{TRD training loss curves for Stable Diffusion 3.5 Medium. EMA smoothing is applied for clearer visualization.}
\label{fig:sd3-losses}
\end{figure}

\begin{figure}[t]
\centering
\includegraphics[width=\linewidth]{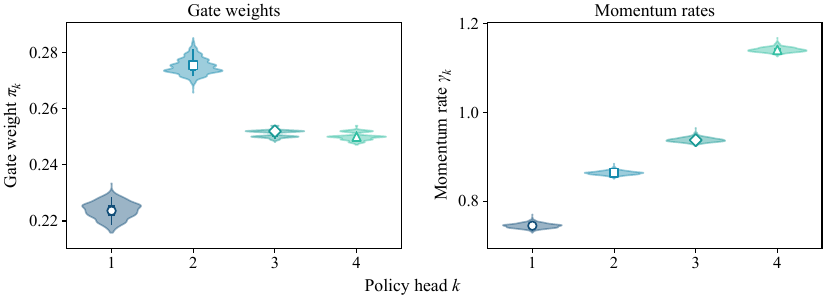}
\caption{Distributions of the gate weights $\pi_k$ and momentum rates $\gamma_k$ produced at $t=1$ by the four policy heads of a TRD checkpoint distilled from Stable Diffusion 3.5 Medium. Appendix~\ref{app:conventions} gives the plotting conventions.}
\label{fig:momentum}
\end{figure}

\begin{figure}[p]
\centering
\includegraphics[width=0.85\linewidth,keepaspectratio]{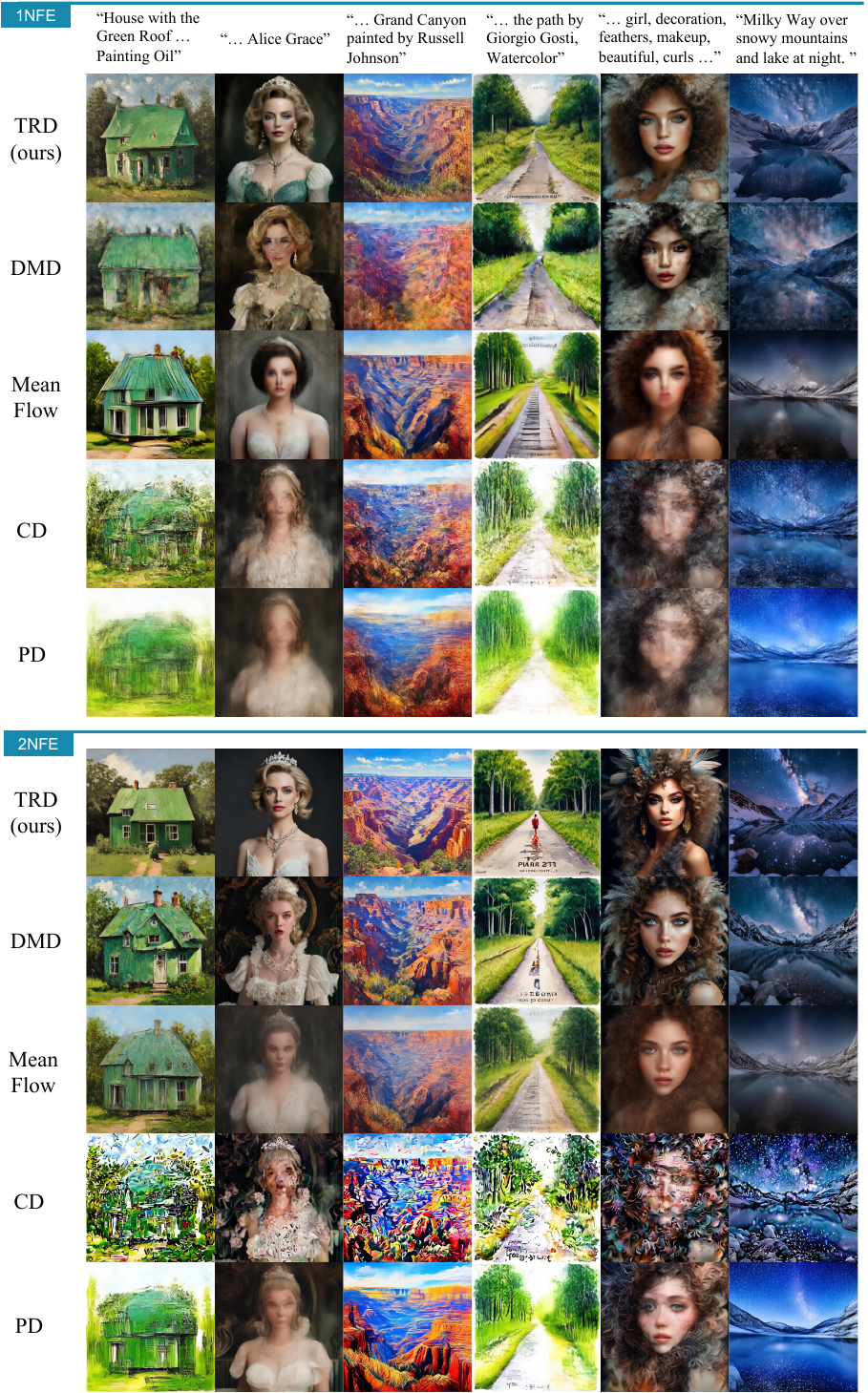}
\caption{Qualitative comparison of PD, CD, MeanFlow, DMD and TRD at one and two steps, distilled from SANA-1.6B.}
\label{fig:qualitative-sana}
\end{figure}

This appendix reports the Stable Diffusion 3.5 Medium comparison in Table~\ref{tab:sd35}, the objective, estimator and policy-capacity ablations in Tables~\ref{tab:loss}, \ref{tab:fake} and~\ref{tab:policy}, the policy-head distribution analysis of \figref{fig:momentum}, the Stable Diffusion 3.5 Medium training-loss curves in \figref{fig:sd3-losses}, and the qualitative results in Figures~\ref{fig:qualitative-sana} to~\ref{fig:additional-sd35}. All quantitative tables use the 10{,}000-image evaluation protocol of Appendix~\ref{app:impl}.

\begin{table}[h]
\caption{One-step and few-step results at $1024\times1024$ for students distilled from Stable Diffusion 3.5 Medium. Conventions follow Table~\ref{tab:sana}.}
\label{tab:sd35}
\begin{center}
\footnotesize
\setlength{\tabcolsep}{4pt}
\fitwidth{%
\begin{tabular}{lcccccccccc}
\toprule
Method & NFE & Extra models & FD-DINOv2 $\downarrow$ & Pick $\uparrow$ & HPS $\uparrow$ & ImgRwd $\uparrow$ & Aesth $\uparrow$ & CLIP $\uparrow$ & Recall $\uparrow$ & Diversity $\uparrow$ \\
\midrule
\textcolor{gray}{SD3.5-Medium teacher} & \textcolor{gray}{$40\times2$} & \textcolor{gray}{--} & \textcolor{gray}{--} & \textcolor{gray}{0.2081} & \textcolor{gray}{0.2425} & \textcolor{gray}{0.8684} & \textcolor{gray}{6.2758} & \textcolor{gray}{32.2167} & \textcolor{gray}{--} & \textcolor{gray}{0.2828} \\
\midrule
\multicolumn{11}{l}{\textit{Trajectory-oriented objectives}} \\
PD & 1 & 0 & 453.30 & 0.1706 & 0.1823 & -2.0211 & 5.0247 & 22.9582 & 0.0080 & 0.2476 \\
CD & 1 & 0 & 385.54 & 0.1737 & 0.1905 & -1.7267 & 5.3324 & 24.3874 & 0.0120 & 0.2194 \\
sCD & 1 & 0 & 270.78 & 0.1713 & 0.1925 & -1.2308 & 5.2646 & 24.8433 & 0.0361 & 0.2202 \\
ArcFlow & 1 & 0 & 293.60 & 0.1773 & 0.1964 & -1.4753 & 5.1572 & 23.9076 & 0.0106 & 0.2317 \\
MeanFlow & 1 & 0 & 177.27 & 0.1808 & 0.2050 & -0.9413 & \underline{5.7254} & 27.1513 & \textbf{0.1479} & \textbf{0.3934} \\
\multicolumn{11}{l}{\textit{Distribution-oriented objectives}} \\
DMD & 1 & 1 & 133.24 & 0.1867 & 0.2116 & -0.1388 & 5.4866 & 29.3045 & 0.0829 & 0.2322 \\
DMD2 & 1 & 2 & \underline{117.75} & 0.1827 & \underline{0.2244} & 0.0090 & 5.5862 & \underline{29.4182} & \underline{0.1355} & 0.2161 \\
SiD & 1 & 1 & 125.33 & \underline{0.1951} & 0.2188 & \underline{0.1253} & 5.4694 & 29.0743 & 0.0960 & 0.2330 \\
FSF-DMD & 1 & 0 & 158.72 & 0.1748 & 0.1829 & -1.3202 & 5.2847 & 25.0290 & 0.0138 & \underline{0.2482} \\
TwinFlow & 1 & 0 & 137.35 & 0.1803 & 0.2128 & -0.2031 & 5.3989 & 28.4420 & 0.0502 & 0.2247 \\
\multicolumn{11}{l}{\textit{Combined trajectory and distribution objective}} \\
TRD (ours) & 1 & 0 & \textbf{86.66} & \textbf{0.2045} & \textbf{0.2433} & \textbf{0.8000} & \textbf{6.2299} & \textbf{31.3551} & 0.1001 & 0.1904 \\
\midrule
\multicolumn{11}{l}{\textit{Trajectory-oriented objectives}} \\
PD & 2 & 0 & 162.17 & 0.1831 & 0.2008 & -1.1153 & 5.6754 & 27.7708 & 0.1630 & \underline{0.3670} \\
CD & 2 & 0 & 161.58 & 0.1866 & 0.2171 & -0.4251 & 5.9082 & 28.4025 & 0.1495 & 0.3383 \\
sCD & 2 & 0 & 161.38 & 0.1825 & 0.2198 & -0.1937 & 5.9318 & 28.2862 & 0.1497 & 0.3293 \\
ArcFlow & 2 & 0 & 176.43 & 0.1840 & 0.2068 & -0.3930 & 5.3749 & 27.3055 & 0.1038 & 0.3442 \\
MeanFlow & 2 & 0 & 159.13 & 0.1815 & 0.2048 & -0.9471 & 5.8877 & 27.0103 & 0.2430 & \textbf{0.4534} \\
\multicolumn{11}{l}{\textit{Distribution-oriented objectives}} \\
DMD & 2 & 1 & \textbf{49.12} & \underline{0.2021} & 0.2327 & \underline{0.6139} & \underline{6.0237} & \underline{31.1482} & \textbf{0.4597} & 0.2375 \\
DMD2 & 2 & 2 & 54.70 & 0.1988 & 0.2334 & 0.5822 & 6.0090 & 31.1444 & \underline{0.3903} & 0.2452 \\
SiD & 2 & 1 & 53.66 & 0.1978 & \underline{0.2344} & 0.5529 & 5.8950 & 29.9768 & 0.3887 & 0.2409 \\
FSF-DMD & 2 & 0 & 68.43 & 0.1958 & 0.2127 & 0.4285 & 5.7737 & 29.6091 & 0.2514 & 0.2212 \\
TwinFlow & 2 & 0 & 64.46 & 0.2001 & 0.2305 & 0.3867 & 5.9101 & 30.0574 & 0.3834 & 0.1962 \\
\multicolumn{11}{l}{\textit{Combined trajectory and distribution objective}} \\
TRD (ours) & 2 & 0 & \underline{49.90} & \textbf{0.2047} & \textbf{0.2435} & \textbf{0.9276} & \textbf{6.1135} & \textbf{31.3486} & 0.2639 & 0.1932 \\
\bottomrule
\end{tabular}
}
\end{center}
\end{table}

\begin{table}[h]
\caption{Contribution of the objective components in \eqref{eq:total} for one-step distillation from SANA-1.6B.}
\label{tab:loss}
\begin{center}
\footnotesize
\setlength{\tabcolsep}{4pt}
\fitwidth{%
\begin{tabular}{lcccccccc}
\toprule
Setting & FD-DINOv2 $\downarrow$ & Pick $\uparrow$ & HPS $\uparrow$ & ImgRwd $\uparrow$ & Aesth $\uparrow$ & CLIP $\uparrow$ & Recall $\uparrow$ & Diversity $\uparrow$ \\
\midrule
w/o $\gL_{\mathrm{tr}}$ & 107.01 & 0.1880 & 0.2179 & 0.3242 & 6.0409 & 28.0689 & 0.1703 & 0.3258 \\
w/o $\gL_{\mathrm{fm}}$ \& $\gL_{\mathrm{dm}}$ & 564.73 & 0.1657 & 0.1821 & -2.0941 & 5.1234 & 20.4366 & 0.0052 & 0.1987 \\
w/o $\gL_{\mathrm{dp}}$ & \multicolumn{8}{c}{\textit{training collapsed}} \\
full objective & 70.34 & 0.1915 & 0.2223 & 0.3888 & 6.1169 & 28.3647 & 0.3162 & 0.3204 \\
\bottomrule
\end{tabular}
}
\end{center}
\end{table}

\begin{table}[h]
\caption{Shared self-estimation against a separately trained fake velocity network for one-step distillation from SANA-1.6B.}
\label{tab:fake}
\begin{center}
\footnotesize
\setlength{\tabcolsep}{3pt}
\fitwidth{%
\begin{tabular}{lccccccccc}
\toprule
Estimator & Extra models & FD-DINOv2 $\downarrow$ & Pick $\uparrow$ & HPS $\uparrow$ & ImgRwd $\uparrow$ & Aesth $\uparrow$ & CLIP $\uparrow$ & Recall $\uparrow$ & Diversity $\uparrow$ \\
\midrule
Shared self-estimator (default) & 0 & 70.34 & 0.1915 & 0.2223 & 0.3888 & 6.1169 & 28.3647 & 0.3162 & 0.3204 \\
Separate fake velocity network & 1 & 127.86 & 0.1861 & 0.2137 & -0.0293 & 5.9505 & 27.7603 & 0.1772 & 0.3337 \\
\bottomrule
\end{tabular}
}
\end{center}
\end{table}

\begin{table}[h]
\caption{Policy capacity for one-step distillation from SANA-1.6B, varying the number of momentum modes $K$ and the number of duplicated final blocks $L$.}
\label{tab:policy}
\begin{center}
\footnotesize
\setlength{\tabcolsep}{4pt}
\fitwidth{%
\begin{tabular}{lccccccccc}
\toprule
Setting & Added trainable params & FD-DINOv2 $\downarrow$ & Pick $\uparrow$ & HPS $\uparrow$ & ImgRwd $\uparrow$ & Aesth $\uparrow$ & CLIP $\uparrow$ & Recall $\uparrow$ & Diversity $\uparrow$ \\
\midrule
$L=1, K=1$ & 5\% & 61.41 & 0.1913 & 0.2228 & 0.4081 & 6.1096 & 28.3633 & 0.3659 & 0.3260 \\
$L=1, K=2$ & 10\% & 53.53 & 0.1936 & 0.2236 & 0.4545 & 6.1845 & 28.3687 & 0.4744 & 0.3406 \\
$L=1, K=4$ & 20\% & 70.34 & 0.1915 & 0.2223 & 0.3888 & 6.1169 & 28.3647 & 0.3162 & 0.3204 \\
$L=1, K=8$ & 40\% & 72.82 & 0.1917 & 0.2220 & 0.3404 & 6.1651 & 28.3131 & 0.3148 & 0.3357 \\
\midrule
$K=4, L=0$ & $\sim$ 0\% & 79.65 & 0.1909 & 0.2206 & 0.3229 & 6.0988 & 28.3210 & 0.2718 & 0.3259 \\
$K=4, L=1$ & 20\% & 70.34 & 0.1915 & 0.2223 & 0.3888 & 6.1169 & 28.3647 & 0.3162 & 0.3204 \\
$K=4, L=2$ & 40\% & 65.24 & 0.1921 & 0.2240 & 0.3982 & 6.1258 & 28.5433 & 0.3467 & 0.3209 \\
\bottomrule
\end{tabular}
}
\end{center}
\end{table}

\begin{figure}[h]
\centering
\includegraphics[width=0.85\linewidth,keepaspectratio]{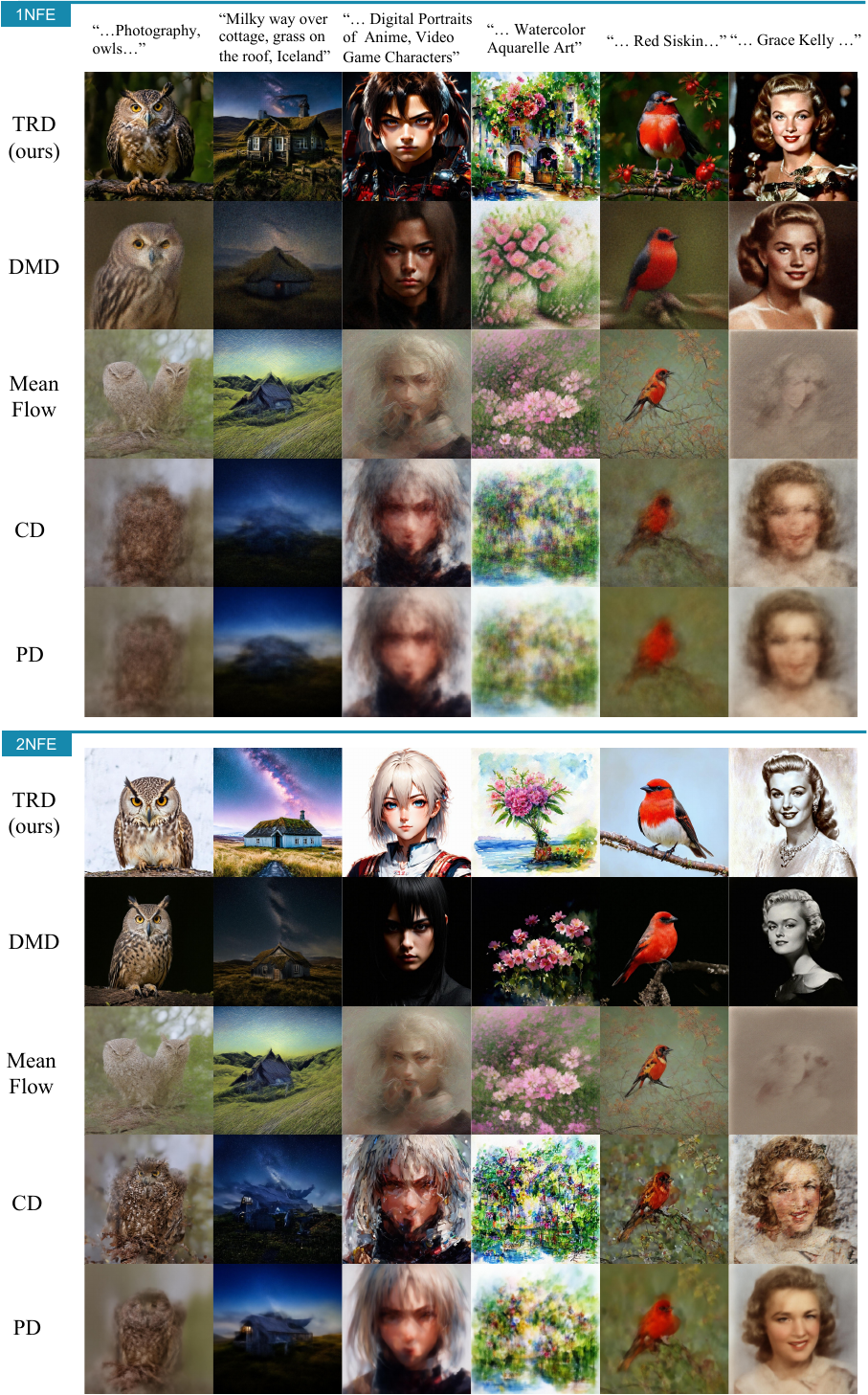}
\caption{Qualitative comparison of PD, CD, MeanFlow, DMD and TRD at one and two steps, distilled from Stable Diffusion 3.5 Medium.}
\label{fig:qualitative-sd35}
\end{figure}

\begin{figure}[p]
\centering
\includegraphics[width=0.92\linewidth]{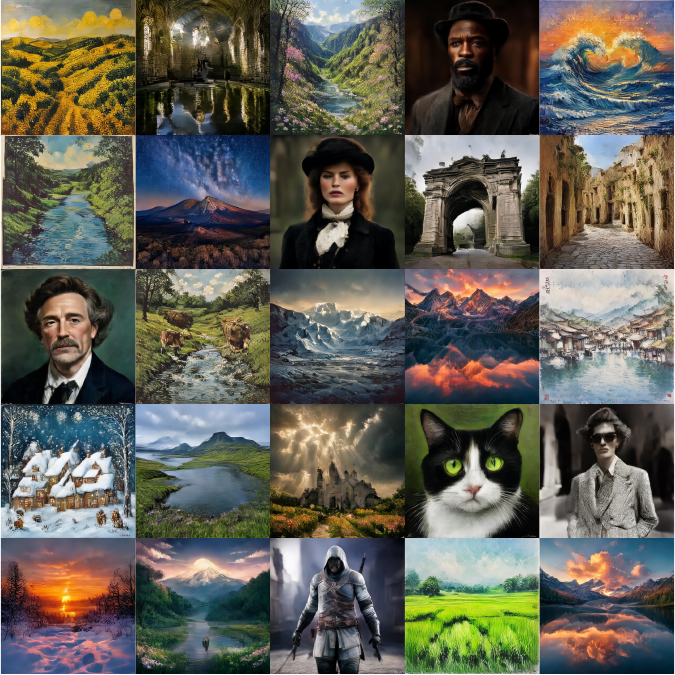}
\caption{Additional one-step TRD samples distilled from SANA-1.6B.}
\label{fig:additional-sana}
\end{figure}

\begin{figure}[p]
\centering
\includegraphics[width=0.92\linewidth]{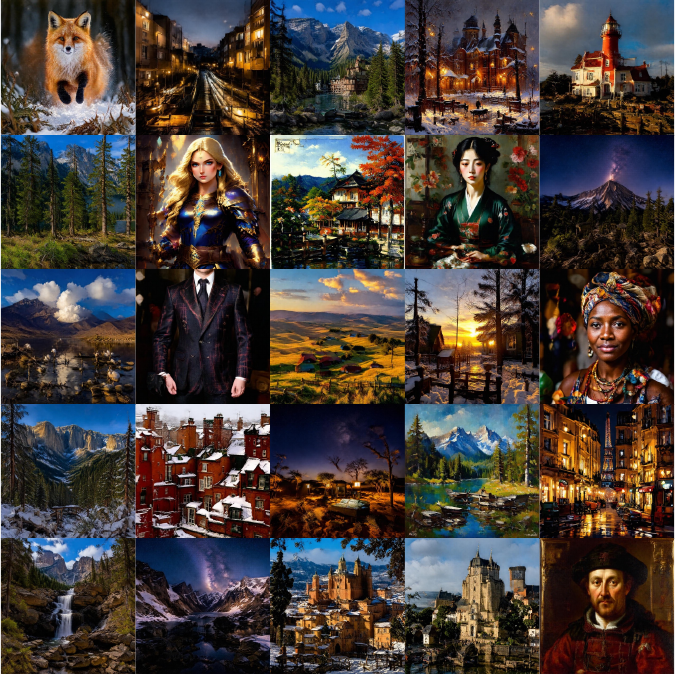}
\caption{Additional one-step TRD samples distilled from Stable Diffusion 3.5 Medium.}
\label{fig:additional-sd35}
\end{figure}

\end{document}